\documentclass[11pt]{article}

\usepackage[utf8]{inputenc} %
\usepackage[T1]{fontenc}    %
\usepackage{hyperref}       %
\usepackage{url}            %
\usepackage{booktabs}       %
\usepackage{amsfonts}       %
\usepackage{nicefrac}       %
\usepackage{microtype}      %
\usepackage{xcolor}         %
\usepackage{color}
\usepackage[margin=1in]{geometry}

\definecolor{myfavblue}{rgb}{0.05, 0.2, 0.8}
\definecolor{keywords}{RGB}{255,0,90}
\definecolor{comments}{RGB}{0,0,113}
\definecolor{red}{RGB}{160,0,0}
\definecolor{green}{RGB}{0,150,0}
\definecolor{C0}{rgb}{0.12156862745098039, 0.4666666666666667, 0.7058823529411765}  %

\definecolor{myblue}{HTML}{3182bd}
\definecolor{myred}{HTML}{de2d26}

\definecolor{mydarkblue}{rgb}{0,0.08,0.45}
\hypersetup{
    colorlinks=true,
    linkcolor=mydarkblue,
    citecolor=mydarkblue,
    filecolor=mydarkblue,
    urlcolor=mydarkblue
}
\usepackage{enumitem}

\usepackage{amsmath,stackrel}
\usepackage{amsthm}
\allowdisplaybreaks
\usepackage{amssymb}
\usepackage{algorithm}
\usepackage{algorithmic}
\usepackage[lined,boxed,ruled,norelsize,algo2e,linesnumbered]{algorithm2e}
\usepackage{subfig}

\usepackage{array}
\usepackage{color}
\usepackage{mathtools}

\usepackage{thmtools} 
\usepackage{thm-restate}

\theoremstyle{plain}
\newtheorem{theorem}{Theorem}[section]
\newtheorem{lemma}[theorem]{Lemma}
\newtheorem{lem}[theorem]{Lemma}

\newtheorem{assumption}[theorem]{Assumption}

\theoremstyle{definition}
\newtheorem{definition}[theorem]{Definition}

\theoremstyle{remark}
\newtheorem{remark}[theorem]{Remark}

\newcommand{\ox}{\overline{x}}

\renewcommand{\epsilon}{\varepsilon}
\renewcommand{\phi}{\varphi}

\newcommand{\defeq}{\stackrel{{\mathrm{def}}}{=}}

\DeclareMathOperator*{\E}{\mathbb{E}}

\DeclareMathOperator*{\argmin}{argmin}

\DeclareMathOperator{\rank}{rank}

\DeclareMathAlphabet{\mathpzc}{OT1}{pzc}{m}{it}

\DeclarePairedDelimiterX{\xdivergence}[2]{(}{)}{%
  #1\;\delimsize\|\;#2%
}

\definecolor{burntorange}{rgb}{0.8, 0.33, 0.0}

\newcommand{\lp}{\left(}

\newcommand{\rp}{\right)}

\newcommand{\R}{\mathbb{R}} %

\newcommand{\cA}{\mathcal A}

\newcommand{\cD}{\mathcal D}

\newcommand{\cK}{\mathcal K}

\newcommand{\cM}{\mathcal M}
\newcommand{\cN}{\mathcal N}
\newcommand{\cO}{\mathcal O}
\newcommand{\cP}{\mathcal P}

\newcommand{\norm}[1]{\left\lVert #1\right\rVert} %

\renewcommand{\epsilon}{\varepsilon}

  \newcommand{\beq}{\begin{equation}}
  \newcommand{\eeq}{\end{equation}}
  \newcommand{\beqn}{\begin{equation*}}
  \newcommand{\eeqn}{\end{equation*}}
  \newcommand{\beqr}{\begin{eqnarray}}
  \newcommand{\eeqr}{\end{eqnarray}}
  \newcommand{\beqrn}{\begin{eqnarray*}}
  \newcommand{\eeqrn}{\end{eqnarray*}}
  \newcommand{\bmline}{\begin{multline}}
  \newcommand{\emline}{\end{multline}}
  \newcommand{\bmlinen}{\begin{multline*}}
  \newcommand{\emlinen}{\end{multline*}}

\newcommand{\Tnabla}{\Tilde{\nabla}}

\definecolor{b2}{RGB}{51,153,255}
\definecolor{mygreen}{RGB}{80,180,0}

\newcommand{\normop}[1]{\left\lVert#1\right\rVert_{\mathrm{op}}}

\newcommand{\normtwo}[1]{\left\lVert#1\right\rVert_2}
\newcommand{\eq}[1]{\begin{align}#1\end{align}}
\newcommand{\eqn}[1]{\begin{align*}#1\end{align*}}
\newcommand*{\bracks}[1]{\left(#1\right)}  %
\newcommand*{\sbracks}[1]{\left[#1\right]}  %

\title{
When Does Differentially Private Learning \\Not Suffer in High Dimensions?
}

\date{\today}

\author{
  \thanks{First two authors contributed equally. Remaining authors listed in alphabetical order.}
  Xuechen Li\thanks{Stanford University, \texttt{lxuechen@cs.stanford.edu}}
  \quad
  Daogao Liu\thanks{University of Washington, \texttt{dgliu@uw.edu}}
  \quad
  Tatsunori Hashimoto\thanks{Stanford University, \texttt{thashim@stanford.edu}}
  \quad
  Huseyin A. Inan\thanks{Microsoft Research,  \texttt{huseyin.inan@microsoft.com}}\\
  \quad
  Janardhan Kulkarni\thanks{Microsoft Research, \texttt{jakul@microsoft.com}}
  \quad
  Yin Tat Lee\thanks{University of Washington and Microsoft Research, \texttt{yintat@uw.edu}}
  \quad
  Abhradeep Guha Thakurta\thanks{Google Research, \texttt{athakurta@google.com}}
}

\begin{document}

\maketitle

\begin{abstract}
  Large pretrained models can be fine-tuned with differential privacy to achieve performance approaching that of non-private models.
  A common theme in these results is the surprising observation that high-dimensional models can achieve favorable privacy-utility trade-offs. 
  This seemingly contradicts known results on the model-size dependence of differentially private convex learning and raises the following research question:
  When does the performance of differentially private learning not degrade with increasing model size? 
  We identify that the magnitudes of gradients projected onto subspaces is a key factor that determines performance.
  To precisely characterize this for private convex learning, we introduce a condition on the objective that we term \emph{restricted Lipschitz continuity} and derive improved bounds for the excess empirical and population risks that are dimension-independent under additional conditions.
  We empirically show that in private fine-tuning of large language models, gradients obtained during fine-tuning are mostly controlled by a few principal components.
  This behavior is similar to conditions under which we obtain dimension-independent bounds in convex settings.
  Our theoretical and empirical results together provide a possible explanation for the recent success of large-scale private fine-tuning.
  Code to reproduce our results can be found at \url{https://github.com/lxuechen/private-transformers/tree/main/examples/classification/spectral_analysis}.
  \end{abstract}

\section{Introduction}

Recent works have shown that large publicly pretrained models can be differentially privately fine-tuned on small downstream datasets with performance approaching those attained by non-private models. 
In particular, past works showed that pretrained BERT~\cite{devlin2018bert} and GPT-2~\cite{radford2018improving,radford2019language} models can be fine-tuned to perform well for text classification and generation under a privacy budget of $\epsilon \in [2, 6]$~\cite{li2021large,yu2021differentially}. 
More recently, it was shown that pretrained ResNets~\cite{he2016deep} and vision-Transformers~\cite{dosovitskiy2020image} can be fine-tuned to perform well for ImageNet classification under single digit privacy budgets~\cite{de2022unlocking,mehta2022large}. 

One key ingredient in these successes has been the use of large pretrained models with millions to billions of parameters. 
These works generally highlighted the importance of two phenomena: (i) large pretrained models tend to experience good privacy-utility trade-offs when fine-tuned, and (ii) the trade-off improves with the improvement of the quality of the pretrained model (correlated with increase in size).
While the power of scale and pretraining have been demonstrated numerous times in non-private deep learning~\cite{kaplan2020scaling}, one common wisdom in private learning had been that large models tend to perform worse.
This intuition was based on (a) results in differentially private convex optimization, most of which predicted that errors would scale proportionally with the dimension of the learning problem in the worst case, and (b) empirical observations that the noise injected to ensure privacy tends to greatly exceed the gradient in magnitude for large models~\cite{yu2021not,gautum14}. 

For instance, consider the problem of differentially private convex \emph{empirical risk minimization} (ERM).
Here, we are given a dataset of $n$ examples $\cD = \{s_j\}_{j=1}^n \in \mathcal{S}^{n}$, a convex set $\cK \subseteq \R^{d}$ (not necessarily bounded), and the goal is to perform the optimization
\eqn{
	\text{minimize}_{x\in \cK} F(x; \cD) = \frac{1}{n} \sum_{j=1}^n f(x; s_j)
}
subject to differential privacy, 
where $f(\cdot;s)$ is convex over $\cK$ for all $s\in \mathcal{S}$. 
For bounded $\cK$, past works presented matching upper and lower bounds that are dimension-dependent under the usual Lipschitz assumption on the objective~\cite{bassily2014private,chaudhuri2011differentially}.
These results seem to suggest that the performance of differentially private ERM algorithms inevitably degrades with increasing problem size in the worst case, and present a seeming discrepancy between recent empirical results on large-scale fine-tuning.\footnote{We judiciously choose to describe the discrepancy as seeming, since the refined analysis presented in the current work suggests that the discrepancy is likely non-existent.}

To better understand the relation between problem size and the performance of differentially private learning, we study the following question both theoretically and empirically:
\begin{quote}
	\centering
	\emph{When does the performance of differentially private stochastic gradient descent (DP-SGD) not degrade with increasing problem dimension?}
\end{quote}

On the theoretical front, we show that DP-SGD can result in dimension-independent error bounds even when gradients span the entire ambient space for unconstrained optimization problems.
We identify that the standard dependence on the dimension of the ambient space can be replaced by the magnitudes of gradients projected onto subspaces of varying dimensions.
We formalize this in a condition that we call \textit{restricted Lipschitz continuity} and derive refined bounds for the excess empirical and population risks for DP-SGD when loss functions obey this condition. 
We show that when the restricted Lipschitz coefficients decay rapidly, both the excess empirical and population risks become dimension-independent.
This extends a previous work which derived rank-dependent bounds for learning generalized linear models in an unconstrained space~\cite{song2021evading}. 

Our theoretical results shed light on the recent success of large-scale differentially private fine-tuning.
We empirically show that gradients of language models during fine-tuning are mostly controlled by a few principal components --- a behavior that is similar to conditions under which we obtain dimension-independent bounds for private convex ERM. 
This provides a possible explanation for the observation that densely fine-tuning with DP-SGD need not necessarily experience much worse performance than sparsely fine-tuning~\cite{li2021large}. Moreover, it suggests that DP-SGD can be adaptive to problems that are effectively low-dimensional (as characterized by restricted Lipschitz continuity) without further algorithmic intervention.

We summarize our contributions below.
\begin{itemize}[leftmargin=7mm]
\setlength\itemsep{0.1em}
	\item [(1)] We introduce a condition on the objective function that we term restricted Lipschitz continuity.
		This condition generalizes the usual Lipschitz continuity notion and gives rise to refined analyses when magnitudes of gradients projected onto diminishing subspaces decay rapidly.
	\item [(2)] Under restricted Lipschitz continuity, we present refined bounds on the excess empirical and population risks for DP-SGD when optimizing convex objectives. These bounds generalize previous dimension-independent results~\cite{song2021evading} and are broadly applicable to cases where gradients are full rank but most coordinates only marginally influence the objective.
	\item [(3)] Our theory sheds light on recent successes of large-scale differentially private fine-tuning of language models.
		We show that gradients obtained through fine-tuning mostly lie in a subspace spanned by a few principal components --- a behavior similar to when optimizing a restricted Lipschitz continuous loss with decaying coefficients.
		These empirical results provide a possible explanation for the recent success of large-scale private fine-tuning.
\end{itemize}

\section{Preliminaries}
We define the notation used throughout this work and state the problems of differentially private empirical risk minimization and differentially private stochastic convex optimization.
Finally, we give a brief recap of differentially private stochastic gradient descent, and existing dimension-dependent and dimension-independent results in the literature.

\paragraph{Notation \& Terminology.} 
For a positive integer $n \in \mathbb{N}_+$, define the shorthand $[n] = \{1, \dots, n\}$.
For a vector $x \in \R^d$, denote its $\ell_2$-norm by $\normtwo{x}$.
Given a symmetric $M \in \R^{d \times d}$, let $\lambda_1(M) \ge \lambda_2(M) \ge \cdots \ge \lambda_d(M)$ denote its eigenvalues.
Given a positive semidefinite matrix $A$, let $\| x \|_A = (x^\top A x)^{1/2}$ denote the induced Mahalanobis norm.
For scalar functions $f$ and $g$, we write $f \lesssim g$ if there exists a positive constant $C$ such that $f(x) \leq C g(x)$ for all input $x$ in the domain.

\subsection{Differentially Private Empirical Risk Minimization and Stochastic Convex Optimization}
Before stating the theoretical problem of interest, we recall the basic concepts of Lipschitz continuity, convexity, and approximate differential privacy.
\begin{definition}[Lipschitz Continuity]
The loss function $h:\cK\to\R$ is $G$-Lipschitz with respect to the $\ell_2$ norm if for all $x,x' \in \cK$, $|f(x)-f(x')|\leq G\|x-x'\|_2$.
\end{definition}

\begin{definition}[Convexity]
The loss function $h:\cK\to\R$ is convex if $h(\alpha x + (1 - \alpha) y) \le \alpha h(x) + (1 - \alpha) h(y)$, for all $\alpha \in [0, 1]$, and $x, y$ in a convex domain $\cK$.
\end{definition}

\begin{definition}[Approximate Differential Privacy~\cite{dwork2014algorithmic}]
A randomized algorithm $\cM$ is $(\epsilon,\delta)$-differentially private if for all neighboring datasets $\cD$ and $\cD'$ that differ by a single record and all sets $\cO\subset\mathrm{range}(\cM)$, the following expression holds
\begin{align*}
    \Pr[\cM(\cD)\in \cO]\le \exp ({\epsilon}) \Pr[\cM(\cD')\in \cO]+\delta.
\end{align*}
\end{definition}
In this work, we study both \emph{differentially private empirical risk minimization} (DP-ERM) for convex objectives and \emph{differentially private stochastic convex optimization} (DP-SCO).

In DP-ERM for convex objectives, we are given a dataset $\cD=\{s_{j}\}_{j\in[n]} \in \mathcal{S}^n$ of $n$ examples. 
Each per-example loss $f(\cdot ; s_j)$ is convex over the convex body $\cK\subseteq\R^{d}$ and $G$-Lipschitz.
We aim to design an $(\epsilon,\delta)$-DP algorithm that returns a solution $x^\text{priv}$ which approximately minimizes the empirical risk $F(x;\cD):=\frac{1}{n}\sum_{s_{j}\in \cD}f(x;s_{j})$.
The term $\mathbb{E}_{x^\text{priv}} \sbracks{ F(x^\text{priv};\cD)-\min_{x\in\cK}F(x;\cD) }$ is referred to as the \emph{excess empirical risk}.

In DP-SCO, we assume the per-example loss $f(\cdot;s)$ is convex and $G$-Lipschitz for all $s\in \mathcal{S}$, and we are given $n$ examples drawn i.i.d. from some (unknown) distribution $\cP$.
The goal is to design an $(\epsilon,\delta)$-DP algorithm which approximately minimizes the population risk
$F(x;\cP):=\E_{s\sim\cP} [ f(x;s) ]$.
The term $\mathbb{E}_{x^\text{priv}} \sbracks{ F(x^\text{priv} ; \cP)-\min_{x\in\cK}F(x;\cP) }$ is referred to as the \emph{excess population risk}.

\subsection{Differentially Private Stochastic Gradient Descent}
\emph{Differentially Private Stochastic Gradient Descent} (DP-SGD)~\cite{abadi2016deep,song2013stochastic} is a popular algorithm for DP convex optimization.
For the theoretical analysis, we study DP-SGD for \emph{unconstrained} optimization. 
To facilitate analysis, we work with the $\ell_{2}$ regularized objective expressed as 
\eqn{
    F_\alpha (x; \cD) = \frac{1}{n} \sum_{j=1}^{n} f(x ; s_j) + \frac{\alpha}{2} \| x - x^{(0)} \|_2^2. 
}
To optimize this objective, DP-SGD independently samples an example in each iteration and updates the solution by combining the gradient with an isotropic Gaussian whose scale is proportional to $G$, the Lipschitz constant of $f$. 
Algorithm \ref{alg:noisy_SGD} presents the pseudocode.
\begin{algorithm2e}
\caption{DP-SGD for optimizing regularized finite-sum objectives}
\label{alg:noisy_SGD}
{\bf Input:} Initial iterate $x^{(0)}$, dataset $\cD = \{s_j\}_{j \in [n]}$, per-step noise magnitude $\sigma$, number of updates $T$, learning rate $\eta$, Lipschitz constant $G$ of $f$. \\
\For{$t=1,\ldots,T$}
{
$j_t \sim \text{Uniform}([n])$\\
$x^{(t)} = 
    x^{(t-1)} -
    \eta\Big(\nabla f(x^{(t-1)};s_{j_t}) + 
    \alpha(x^{(t-1)}-x^{(0)})+ G \cdot \zeta_t \Big), \quad \zeta_t \sim \cN(0,\sigma^{2}I_d)$
}
{\bf Return:} $\overline{x}\defeq\frac{1}{T}\sum_{t=1}^{T}x^{(t)}$.
\end{algorithm2e}

It is straightforward to show that Algorithm~\ref{alg:noisy_SGD} satisfies differential privacy. 
The following lemma quantifies the overall privacy spending and builds on a long line of work on accounting the privacy loss of compositions~\cite{abadi2016deep,balle2018privacy}.
\begin{lemma}[{\cite{KLL21}}]
\label{lm:privacy_guarantee}
There exists constants $c_1$ and $c_2$ such that for $n\geq10$, $\epsilon< c_1 T/n^2$ and $\delta \in (0, \frac{1}{2}]$, DP-SGD (Algorithm~\ref{alg:noisy_SGD}) is $(\epsilon,\delta)$-DP whenever $\sigma \ge \frac{c_2 \sqrt{T\log(1/\delta)}}{\epsilon n}$.
\end{lemma}

\subsection{On the Dimension Dependence of Private Learning}
Early works on bounding the excess empirical and population risks for privately optimizing convex objectives focused on a constrained optimization setup where algorithms typically project iterates back onto a fixed bounded domain after each noisy gradient update. 
Results in this setting suggested that risks are inevitably dimension-dependent in the worst case.
For instance, it was shown that the excess empirical risk bound $\Theta (  G \normtwo{\cK} \sqrt{d\log(1/\delta)} n^{-1} \epsilon^{-1} )$ and excess population risk bound $\Theta ( G \normtwo{\cK} ( n^{-1/2} + \sqrt {d \log(1/\delta) } n^{-1} \epsilon^{-1} ) )$ are tight when privately optimizing convex $G$-Lipschitz objectives in a convex domain of diameter $\normtwo{\cK}$~\cite{bassily2014private}.
Moreover, the lower bound instances in these works imply that such dimension-dependent lower bounds also apply when one considers the class of loss functions whose gradients are low-rank.

The body of work on unconstrained convex optimization is arguably less abundant, with the notable result that differentially private gradient descent need not suffer from a dimension-dependent penalty when learning generalized linear models with low-rank data (equivalently stated, when gradients are low-rank)~\cite{song2021evading}.
Our main theoretical results (Theorems~\ref{thm:DPERM} and~\ref{thm:DPSCO}) extend this line of work and show that dimension-independence is achievable under weaker conditions.

\section{Dimension-Independence via Restricted Lipschitz Continuity}\label{sec:theory}
In this section, we introduce the restricted Lipschitz continuity condition and derive improved bounds for the excess empirical and population risks for DP-SGD when optimizing convex objectives.

\begin{definition}[Restricted Lipschitz Continuity]
We say that the loss function $h: \cK \to \R$ is restricted Lipschitz continuous with coefficients $\{G_k\}_{k=0}^d$, if
for all $k \in \{0, \dots, d\}$, there exists an orthogonal projection matrix $P_k$ with rank $k$ such that 
\begin{align*}
    \|(I-P_k)\nabla h(x) \|_2\le G_k,
\end{align*}
for all $x \in \cK$ and all subgradients $\nabla h(x) \in \partial h(x)$.
\end{definition}

Note that any $G$-Lipschitz function is also trivially restricted Lipschitz continuous with coefficients $G = G_0 = G_1 = \cdots = G_d$, since orthogonal projections never increase the $\ell_2$-norm of a vector (generally, it is easy to see that $G = G_{0}\ge G_{1}\geq G_{2}\geq\cdots\geq G_{d}=0$). 
On the other hand, we expect that a function which exhibits little growth in some subspace of dimension $k$ to have a restricted Lipschitz coefficient $G_{d-k}$ of almost 0.
Our bounds on DP convex optimization characterize errors through the use of restricted Lipschitz coefficients. 
We now summarize the main assumption.
\begin{assumption}
\label{assm:G_k}
The per-example loss function $f(\cdot; s)$ is convex and $G$-Lipschitz continuous for all $s \in \mathcal{S}$.
The empirical loss $F(\cdot; \cD)$ is restricted Lipschitz continuous with coefficients $\{G_k\}_{k = 0}^d$.
\end{assumption}

\subsection{Bounds for Excess Empirical Loss}
We present the main theoretical result on DP-ERM for convex objectives. 
The result consists of two components:
Equation~\eqref{eq:erm_general_condition} is a general bound that is applicable to any sequence of restricted Lipschitz coefficients;
Equation~\eqref{eq:erm_fast_decaying} specializes the previous bound when the sequence of coefficients decays rapidly and demonstrates dimension-independent error scaling.
\begin{restatable}[Excess Empirical Loss]{theorem}{DPERM}
\label{thm:DPERM}
Let $ \delta \in (0, \frac{1}{2}]$ and $\epsilon\in(0,10]$.
Under Assumption~\ref{assm:G_k}, for all $k \in [d]$, 
setting 
$T=\Theta(n^2 + d \log^2 d)$, $\sigma=\Theta \bracks{ \frac{\sqrt{T\log(1/\delta)}}{n\epsilon} }$, $\eta=\sqrt{\frac{D^{2}}{T\cdot G_{0}^{2}\cdot k\sigma^{2}}}$
and $\alpha=\frac{1}{D}\sqrt{\sum_{s=1}^S s^2 2^{s}G_{2^{s-1}k}^{2}}$, where $S = \lfloor \log (d / k) \rfloor + 1$, DP-SGD (Algorithm~\ref{alg:noisy_SGD}) is $(\epsilon,\delta)$-DP, and
\begin{align}
\label{eq:erm_general_condition}
\E \sbracks{  F(\overline{x};\cD)-  \min_x F(x;\cD) } &~ \lesssim \frac{G_{0}D\sqrt{k\log(1/\delta)}}{\epsilon n}+D\sqrt{\sum_{s=1}^{S}s^2 2^{s}G_{2^{s-1}k}^{2}},
\end{align}
where $\|x^{(0)}- \argmin_x F(x; \cD) \|_2 \le D$, $\ox$ is the (random) output of DP-SGD (Algorithm~\ref{alg:noisy_SGD}), and the expectation is over the randomness of $\overline{x}$.
Moreover, if for some $c>1/2$, we have $G_k\leq G_0 k^{-c}$ for all $k \in [d]$, and in addition $ n\geq \epsilon^{-1} \sqrt{\log(1/\delta)}$,
minimizing the right hand side of \eqref{eq:erm_general_condition} with respect to $k$ yields
\begin{align}
\label{eq:erm_fast_decaying}
\E \sbracks{  F(\overline{x};\cD)-  \min_x F(x;\cD) } \lesssim G_{0}D \cdot \bracks{ 
        \frac{\sqrt{\log(1/\delta)}}{\epsilon n}
    }^{2c/(1+2c)
}.
\end{align}
\end{restatable}
We include a sketch of the proof techniques in Subsection~\ref{sec:proof_technique} and defer the full proof to Subsection~\ref{app:dperm_proof}.
\begin{remark}
Consider DP-ERM for learning generalized linear models with convex and Lipschitz losses.
When the (empirical) data covariance is of rank $r < d$, the span of gradients $\text{span}(\{ \nabla_x F(x) \})$ is also of rank $r$. 
Thus, the average loss is restricted Lipschitz continuous with coefficients where $G_{r'} = 0$ for all $r' > r$. 
Setting $k = r$ in \eqref{eq:erm_general_condition} yields an excess empirical risk bound of order $O\bracks{ {G_{0}D\sqrt{ r \cdot\log(1/\delta)}} \epsilon^{-1} n^{-1} }$.
This recovers the previous dimension-independent result in~\cite{song2021evading}.
\end{remark}

The restricted Lipschitz continuity condition can be viewed as a generalized notion of rank.
The result captured in \eqref{eq:erm_fast_decaying} suggests that the empirical loss achieved by DP-SGD does not depend on the problem dimension if the sequence of restricted Lipschitz coefficients decays rapidly. 
We leverage these insights to build intuition for understanding privately fine-tuning language models in Section~\ref{sec:experiments}.

\subsection{Bounds for Excess Population Loss}
For DP-SCO, we make use of the \emph{stability} of DP-SGD to bound its generalization error \cite{BE02}, following previous works \cite{bassily2019private,BFGT20,song2021evading}.
The bound on the excess population loss follows from combining the bounds on the excess empirical risk and the generalization error.

\begin{restatable}[Excess Population Loss]{theorem}{DPSCO}
\label{thm:DPSCO}
Let $ \delta \in (0, \frac{1}{2}]$ and $\epsilon\in (0,10]$.
Under Assumption~\ref{assm:G_k}, for all $k \in [d]$, 
by setting
$T=\Theta(n^2 + d \log^2 d)$,
$\sigma=\Theta \bracks{ \frac{\sqrt{T\log(1/\delta)}}{n\epsilon} }$, 
$\eta=\sqrt{\frac{D^{2}}{T\cdot G_{0}^{2}(T/n+k\sigma^{2})}}$
and $\alpha=\frac{1}{D}\sqrt{\sum_{s=1}^S s^2 2^{s}G_{2^{s-1}k}^{2}}$, where $S = \lfloor \log(d/k) \rfloor + 1$, 
DP-SGD (Algorithm~\ref{alg:noisy_SGD}) is $(\epsilon,\delta)$-DP, and
\begin{align*}
\E \sbracks{
    F(\ox; \cP) - \min_{x} F(x; \cP)
}
\lesssim
    \frac{G_0D}{\sqrt{n}}+\frac{G_{0}D\sqrt{k\log(1/\delta)}}{\epsilon n}+D\sqrt{\sum_{s=1}^{S}s^2 2^{s}G_{2^{s-1}k}^{2}},
\end{align*}
where $\|x^{(0)}- \argmin_x F(x ; \cP) \|_2 \le D$, $\ox$ is the (random) output of DP-SGD (Algorithm~\ref{alg:noisy_SGD}), and the expectation is over the randomness of $\overline{x}$.

Moreover, if for some $c>1/2$, we have $G_k\leq G_0 k^{-c}$ for all $k \in [d]$, and in addition $n>\epsilon^{-1}\sqrt{\log(1/\delta)}$, minimizing the above bound with respect to $k$ yields
\begin{align*}
\E \sbracks{
    F(\ox;\cP)- \min_x F(x;\cP)
}
\lesssim
    \frac{G_0D}{\sqrt{n}}+
    G_{0} D \cdot \bracks{
        \frac{\sqrt{\log(1/\delta)}}{\epsilon n}
    }^{2c/(1+2c)}.
\end{align*}
\end{restatable}

\begin{remark}
Our result on DP-SCO also recovers the DP-SCO rank-dependent result for learning generalized linear models with convex and Lipschitz losses~\cite{song2021evading}.
\end{remark}

\begin{remark}
When $c>1/2$, $\epsilon = \Theta(1)$ and $\delta = 1/\mathrm{poly}(n)$, the population loss matches the (non-private) informational-theoretical lower bound $\Omega(G_0 D/\sqrt{n})$ \cite{agarwal2009information}.
\end{remark}

\begin{remark}
    Our results on DP-ERM and DP-SCO naturally generalize to (full-batch) DP-GD.
\end{remark}

\subsection{Overview of Proof Techniques}\label{sec:proof_technique}
The privacy guarantees in Theorems \ref{thm:DPERM} and \ref{thm:DPSCO} follow from Lemma~\ref{lm:privacy_guarantee}.
It suffices to prove the utility guarantees. 
We give an outline of the main proof techniques first and present full proofs afterwards.
The following is a sketch of the core technique for deriving~\eqref{eq:erm_fast_decaying} in Theorem~\ref{thm:DPERM}.
For simplicity, we write $f_j(\cdot)$ for $f(\cdot;s_j)$ and $F(\cdot)$ for $F(\cdot;\cD)$ when there is no ambiguity.

By convexity, the error term of SGD is upper bounded as follows
\eq{
\label{eq:proof_sketch}
    f_{j}(x^{(t)})-f_{j}(x^{*})\leq \nabla f_{j}(x^{(t)})^{\top}(x^{(t)}-x^{*}),
}
where $j \in [n]$ is the random index sampled at iteration $t$.

By definition of $G_{k}$, we know that there is a $k$ dimensional
subspace $U$ such that the gradient component orthogonal to $U$ is small when $G_k$ is small.
Na\"ively, one decomposes the gradient $\nabla f_{j}(x^{(t)})=g_{1}+g_{2}$,
where $g_{1}\in U$ and $g_{2}\in U^{\perp}$,
and separately bounds the two terms $g_1^\top (x^{(t)}-x^{*})$ and $g_2^\top (x^{(t)}-x^{*})$.
Since $g_{1}$ lies in a $k$ dimensional subspace, one can follow existing
arguments on DP-SGD to bound $g_{1}^{\top}(x^{(t)}-x^{*})$. 
Unfortunately, this argument does not give a dimension-independent bound.
Although
$\|g_{2}\|_{2}\leq G_{k}$ (which can be small for large $k$), the term
$\|x^{(t)}-x^{*}\|_{2}$ is as large as $\Omega(\sqrt{d})$ with high probability due
to the isotropic Gaussian noise injected in DP-SGD.

Our key idea is to partition the whole space $\R^d$ into  $\lfloor \log(d/k) \rfloor + 2$ orthogonal subspaces, expressing the error term $\nabla f_j(x^{(t)})^\top (x^{(t)}-x^*)$ as the sum of individual terms, each of which corresponds to a projection to a particular subspace.
Fix a $k \in [d]$, and consider the following subspaces:
Let $U_{0}=\text{range}(P_{k})$,
 $U_{s}$ be the subspace orthogonal to all previous subspaces such that $\bigoplus_{i=0}^s U_i = \text{range}(P_{2^sk})$ for $s=1,2,\cdots,\left\lfloor \log(d/k)\right\rfloor $,
and $U_{S}$ be the subspace such that the orthogonal direct sum of all subspaces $\{U_i\}_{i=0}^S$ is $\R^d$, where $S=\left\lfloor \log(d/k)\right\rfloor +1$.
Here, $P_{i}$ is the orthogonal projection matrix with rank $i$ promised by $G_{i}$ in Assumption~\ref{assm:G_k}. 
Let $Q_{s}$ be the orthogonal projection to the subspace $U_{s}$, and observe that $\rank (Q_{s}) \leq 2^{s}k$ and
$\|Q_{s}\nabla F(x)\|_{2}\leq G_{2^{s-1}k}\text{ for all }x\text{ and all }s\geq1.$
Rewriting the right hand side of~\eqref{eq:proof_sketch} with this decomposition yields
\eqn{
f_{j}(x^{(t)})-f_{j}(x^{*})  \leq \lp Q_{0}\nabla f_{j}(x^{(t)})+\sum_{s=1}^{S} Q_{s}\nabla f_{j}(x^{(t)})\rp^{\top}(x^{(t)}-x^{*}).    
}
On the one hand, if $G_k$ decays quickly, $\| \mathbb{E}_j [  Q_{s} \nabla f_{j} ] \|_2$ can be small for large $s$. 
On the other hand, we expect $\|Q_{s} (x^{(t)}-x^{*})\|_2$ to be small for small $s$ where $Q_{s}$ is an orthogonal projection onto a small subspace. 
Thus, for each $s$, $\nabla f_{j}(x^{(t)})^{\top} Q_{s} (x^{(t)}-x^{*})$ is small either due to a small gradient (small $Q_s \nabla f_{j}$ in expectation over the random index) or small noise (small $Q_{s} (x^{(t)}-x^{*})$), since noise injected in DP-SGD is isotropic. 
More formally, in Lemma~\ref{lem:radius}, we show that for any projection matrix $Q$ with rank $r$, $\|Q(x^{(t)}-x^{(0)})\|_2$ can be upper bounded by a term that depends only on $r$ (rather than $d$).

\subsection{Proof of Theorem~\ref{thm:DPERM}}\label{app:dperm_proof}

Before bounding the utility of DP-SGD, we first bound $x^{(t)}-x^{(0)}$ in expectation.

\begin{lemma}
\label{lem:radius}
Suppose Assumption~\ref{assm:G_k} holds.
Let $Q$ be an orthogonal projection matrix with rank $r$ and suppose that $\|Q\nabla f(x;s)\|_2 \leq G_Q$ for all $x\in\R^d$ and $s \in \mathcal{S}$. 
If we set $\eta\leq\frac{1}{2\alpha}$ in DP-SGD, then for all $t>0$, we have
\[
\E\|Q(x^{(t)}-x^{(0)})\|_2^{2}\leq\frac{4 G_{Q}^{2}}{\alpha^{2}}+\frac{2\eta G_{0}^{2}}{\alpha}(1+r\sigma^{2}).
\]
\end{lemma}

\begin{proof}[Proof of Lemma \ref{lem:radius}]
By the assumption, we know $\|Q \nabla F(x)\|_2\le G_Q$ and $\| \nabla F(x)\|_2\leq G_0$.
Let $z^{(t)}=x^{(t)}-x^{(0)}$. Note that
\begin{align*}
    z^{(t+1)}=& ~ x^{(t+1)}-x^{(0)}\\
    =& ~ x^{(t)}-\eta\Big(\nabla f_j(x^{(t)})+\alpha(x^{(t)}-x^{(0)})+G_{0}\cdot \zeta\Big)-x^{(0)}\\
    =& ~ (1-\alpha\eta)z^{(t)}-\eta(\nabla f_{j}(x^{(t)})+G_{0}\cdot\zeta ),
\end{align*}
where $\zeta\sim \cN(0,\sigma^{2}I_d)$ is the isotropic Gaussian noise drawn in $(t+1)$th step. 
For simplicity, we use $\Tnabla f_j(x^{(t)})$ to denote the noisy subgradient $\nabla f_{j}(x^{(t)})+G_{0}\cdot\zeta$.
Hence, we have
\begin{align*}
    \|Qz^{(t+1)}\|_2^2=& ~ (1-\alpha\eta)^2\|Qz^{(t)}\|_2^2-2\eta(1-\alpha\eta)( Qz^{(t)})^\top Q\Tnabla f_j(x^{(t)})+\eta^2\|Q\Tnabla f_j(x^{(t)})\|_2^2.
\end{align*}

Taking expectation over the random sample $f_j$ and random Gaussian noise $\zeta$, we have
\begin{align*}
\E\|Qz^{(t+1)}\|_{2}^{2}= & (1-\alpha\eta)^{2}\E\|Qz^{(t)}\|_{2}^{2}-2\eta(1-\alpha\eta)\cdot\E\Big((Qz^{(t)})^{\top}(Q\nabla F(x^{(t)}))\Big) \\
 & +\eta^{2}\E\|Q\Tnabla f_j(x^{(t)})\|_{2}^{2}\\
\leq & (1-\alpha\eta)\E[\|Qz^{(t)}\|_{2}^{2}]+2 \eta G_{Q}\cdot\E[\|Qz^{(t)}\|_{2}]+\eta^{2}G_{0}^{2}(1+r\sigma^{2}),
\end{align*}
where we used the fact that $\zeta$ has zero mean, $\|\nabla f_{j}(x^{(t)})\|_{2}\leq G_{0}$, $\|Q\nabla F(x^{(t)})\|_{2}\leq G_{Q}$ and $\eta\leq\frac{1}{2\alpha}$. 
Further simplifying and taking expectation over all iterations, we have
\begin{align}
\label{eq:recursion_distance}
\E\|Qz^{(t+1)}\|_{2}^{2} & \leq(1-\alpha\eta)\E\|Qz^{(t)}\|_{2}^{2}+2\eta(\frac{\alpha}{4}\E\|Qz^{(t)}\|_{2}^{2}+\frac{1}{\alpha}G_{Q}^{2})+\eta^{2}G_{0}^{2}(1+r\sigma^{2}) \nonumber\\
 & \leq(1-\frac{\alpha\eta}{2})\E\|Qz^{(t)}\|_{2}^{2}+\frac{2 \eta}{\alpha}G_{Q}^{2}+\eta^{2}G_{0}^{2}(1+r\sigma^{2}).
\end{align}
Using that $z^{(0)}=0$, we know $\E\|Qz^{(0)}\|_2^2=0$.
Solving the recursion (Equation~\eqref{eq:recursion_distance}) gives 
\[
\E\|Qz^{(t)}\|_{2}^{2}\leq\frac{2}{\alpha\eta}(\frac{2 \eta}{\alpha}G_{Q}^{2}+\eta^{2}G_{0}^{2}(1+r\sigma^{2}))
\]
for all $t$. 
This concludes the proof.
\end{proof}

Now, we are ready to bound the utility. 
The proof builds upon the standard mirror descent proof.
\begin{lem}
\label{lm:utility_DPSGD_appendix}
Let $\delta \in (0, \frac{1}{2}]$, and $\epsilon \in (0, 10]$.
Under Assumption~\ref{assm:G_k}, let $x^{(0)}$ be the initial iterate and $x^{*} \in \R^d$ be such that $\|x^{(0)}-x^{*}\|_2\le D$.
For all $k \in [d]$,
setting $T=\Theta(n^2 + d \log^2 d)$,
 $\sigma=\Theta \bracks{ \frac{\sqrt{T\log(1/\delta)}}{n\epsilon} }$, $\eta=\sqrt{\frac{D^{2}}{T\cdot G_{0}^{2}\cdot k\sigma^{2}}}$
and $\alpha=\frac{1}{D}\sqrt{\sum_{s=1}^S s^2 2^{s}G_{2^{s-1}k}^{2}}$, we have
\begin{align*}
\E [F(\overline{x})- F(x^{*})] &~ \lesssim \frac{G_{0}D\sqrt{k\log(1/\delta)}}{\epsilon n}+D\sqrt{\sum_{s=1}^{S} s^2 2^{s}G_{2^{s-1}k}^{2}},
\end{align*}
where $S = \lfloor \log(d/k) \rfloor + 1$, $\overline{x}$ is the output of DP-SGD, and the expectation is under the randomness of DP-SGD.

Moreover, if $G_k\leq G_0 k^{-c}$ for each $k$ for some $c>1/2$, and in addition $n > \epsilon^{-1} \sqrt{\log(1/\delta)} $, picking the best $k \in [d]$ for the bound above gives
\begin{align*}
    \E[F(\ox;\cD)-F(x^*;\cD)]\lesssim
        G_{0}D\cdot \bracks{
            \frac{\sqrt{\log(1/\delta)}}{\epsilon n}
        }^{2c/(1+2c)}.
\end{align*}
\end{lem}

\begin{proof}[Proof of Lemma \ref{lm:utility_DPSGD_appendix}]
The above statement is true for $k = d$ by standard arguments in past work~\cite{bassily2014private,song2021evading}.
Now fix a $k \in \{1, \dots, d - 1\}$.
Our key idea is to split the whole space $\R^d$ into different subspaces.
We define the following set of subspaces:
\begin{itemize}
\item $U_{0}=\text{range}(P_{k})$.
\item For $s=1,2,\ldots,\left\lfloor \log(d/k)\right\rfloor$, 
    let $U_{s} \subseteq \text{range}(P_{2^{s}k})$ be a subspace with maximal dimension such that $U_s \bot U_i$ for all $i=0, \dots, s-1$.
\item For $S=\left\lfloor \log(d/k)\right\rfloor +1$, 
    let $U_{S} \subseteq \R^{d}$ be the subspace such that $\bigoplus_{i=0}^S U_i = \R^d$, and $U_S \bot U_i$ for all $i=0, \dots, S-1$.
\end{itemize}
Recall $P_{i}$ is the orthogonal projection matrix with rank $i$ that gives rise to $G_{i}$ in Assumption~\ref{assm:G_k}. 
In the above, we have assumed that the base of $\log$ is 2.
Let $Q_{s}$ be the orthogonal projection matrix that projects vectors onto the subspace $U_{s}$. 
Note that $\rank(Q_{s})\leq 2^{s}k$ since $U_s \subseteq \text{range}(P_{2^s k})$.
Moreover, it's clear that $U_s \bot \text{range}(P_{2^{s-1}k})$ for all $s \in \{1, \dots, S\}$. 
This is true by construction $\bigoplus_{i=0}^{s-1} U_i \supseteq \text{range}(P_{2^{s-1}k})$ and that 
$U_s \bot \bigoplus_{i=0}^{s-1} U_i$. Thus,
\begin{equation}
\label{eq:Qbound}
\|Q_{s}\nabla F(x)\|_{2} =
\|Q_{s} (I - P_{2^{s-1}k}) \nabla F(x)\|_{2}
\le 
\normop{Q_{s}} \normtwo{(I - P_{2^{s-1}k}) \nabla F(x)}
\leq G_{2^{s-1}k}
\end{equation}
for all $x \in \R^d$ and all $s \in \{1, \dots, S\}$.

Let $j \in [n]$ be the (uniformly random) index sampled in iteration $t$ of DP-SGD.
By convexity of the individual loss $f_j$,
\begin{align*}
    f_{j}(x^{(t)})-f_{j}(x^{*})\leq\nabla f_{j}(x^{(t)})^{\top}(x^{(t)}-x^{*}).
\end{align*}
By construction, $\R^d$ is the orthogonal direct sum of the subspaces $\{U_j\}_{j=0}^S$, and thus any vector $v\in \R^d$ can be rewritten as the sum $\sum_{i=0}^S Q_i v $.
We thus split the right hand side of the above as follows
\begin{align}
f_{j}(x^{(t)})-f_{j}(x^{*}) & \leq \lp Q_{0}\nabla f_{j}(x^{(t)})+\sum_{s=1}^{S}Q_{s}\nabla f_{j}(x^{(t)})\rp^{\top}(x^{(t)}-x^{*}).\label{eq:mirror_two_term}
\end{align}

We use different approaches to bound $(Q_0\nabla f_j(x^{(t)}))^\top (x^{(t)}-x^*)$ and $(Q_s\nabla f_j(x^{(t)}))^\top (x^{(t)}-x^*)$ when $s\geq 1$, and we discuss them separately in the following.

\medskip
{\bf  Bounding  $(Q_0\nabla f_j(x^{(t)}))^\top (x^{(t)}-x^*)$:} Recall that
\begin{align*}
    x^{(t+1)}=x^{(t)}-\eta\Big(\nabla f_j(x^{(t)})+\alpha(x^{(t)}-x^{(0)})+G_{0}\cdot \zeta\Big)
\end{align*}
for some Gaussian $\zeta\sim \cN(0,\sigma^{2}I_d)$. 
Hence, we have
\begin{align}
\label{eq:duality_Q_0}
    (\nabla f_{j}(x^{(t)}))^{\top}Q_{0}(x^{(t)}-x^{*}) = &~~ \left(\frac{1}{\eta}(x^{(t)}-x^{(t+1)})-\alpha(x^{(t)}-x^{(0)})-G_{0}\cdot\zeta\right)^{\top}Q_{0}(x^{(t)}-x^{*})\nonumber \\
= & ~~ \left(\frac{1}{\eta}Q_0(x^{(t)}-x^{(t+1)})\right)^{\top}Q_{0}(x^{(t)}-x^{*})-\left(
\alpha(x^{(t)}-x^{(0)})+G_{0}\cdot\zeta
\right)^\top Q_0(x^{(t)}-x^*)\nonumber \\
=& \frac{1}{2\eta}
    \bracks{
        \|Q_{0}(x^{(t)}-x^{*})\|_2^{2} - 
        \|Q_{0}(x^{(t+1)}-x^{*})\|_2^{2} + 
        \|Q_{0}(x^{(t)}-x^{(t+1)})\|_2^{2}
    }
\nonumber \\
 ~& -\lp\alpha(x^{(t)}-x^{(0)})+G_{0}\cdot\zeta\rp^{\top}Q_{0}(x^{(t)}-x^{*}),
\end{align}
where we used the fact that $Q_0^2 v = Q_0 v$ for any $v \in \R^d$ (since $Q_0$ is a projection matrix), and the last equality follows from 
\begin{align*}
   2(Q_0(x^{(t)}-x^{(t+1)}))^\top Q_0(x^{(t)}-x^*) \\
 = & \|Q_{0}(x^{(t)}-x^{*})\|_2^{2} - 
    \|Q_{0}(x^{(t+1)}-x^{*})\|_2^{2} + 
    \|Q_{0}(x^{(t)}-x^{(t+1)})\|_2^{2}.
\end{align*}

Taking expectation on $\zeta$ over both sides of Equation~\eqref{eq:duality_Q_0} and making use of the fact that $\zeta$ has mean $0$, we have
\begin{align*}
\E_{\zeta}(Q_{0}\nabla f_{j}(x^{(t)}))^{\top}(x^{(t)}-x^{*}) = & \frac{1}{2\eta}
    \lp\E_{\zeta}\|Q_{0}(x^{(t)}-x^{*})\|_2^{2} - 
        \E_{\zeta}\|Q_{0}(x^{(t+1)}-x^{*})\|_2^{2} + 
        \E_{\zeta}\|Q_{0}(x^{(t)}-x^{(t+1)})\|_2^{2}\rp\\
 & -\alpha\E_{\zeta}\lp (x^{(t)}-x^{(0)})^{\top}Q_{0}(x^{(t)}-x^{*}) \rp.
\end{align*}

Recalling the definition of $Q_0$ and that $Q_0$ has rank at most $k$, one has
\begin{align*}
\E_{\zeta}\|Q_{0}(x^{(t)}-x^{(t+1)})\|_2^{2} =
    &~ \eta^{2}\E_{\zeta}\|Q_{0}\big(\nabla f_{j}(x^{(t)})+\alpha(x^{(t)}-x^{(0)})+G_{0}\cdot\zeta\big)\|_2^{2}\\
    = &~ \eta^{2}\E_{\zeta}\|Q_{0}\big(\nabla f_{j}(x^{(t)})+\alpha(x^{(t)}-x^{(0)})\big)\|_2^{2}+\eta^{2}G_{0}^{2}k\sigma^{2}\\
\leq &~
    2\eta^{2}G_{0}^{2}(1+k\sigma^{2}) + 
    2\eta^{2}\alpha^{2}\E_{\zeta}\|Q_{0}(x^{(t)}-x^{(0)})\|_2^{2}.
\end{align*}

Moreover, one has
\begin{align*}
-\alpha(x^{(t)}-x^{(0)})^{\top}Q_{0}(x^{(t)}-x^{*}) = &~ -\alpha(x^{(t)}-x^{(0)})^{\top}Q_{0}(x^{(t)}-x^{(0)})-\alpha(x^{(t)}-x^{(0)})^{\top}Q_{0}(x^{(0)}-x^{*})\\
\leq &~
    -\frac{\alpha}{2}\|Q_{0}(x^{(t)}-x^{(0)})\|_2^{2} + 
    \frac{\alpha}{2}\|Q_{0}(x^{(0)}-x^{*})\|_2^{2}.
\end{align*}

Therefore, we have
\begin{align}
\E_{\zeta}(Q_{0}\nabla f_{j}(x^{(t)}))^{\top}(x^{(t)}-x^{*}) \leq & ~ \frac{1}{2\eta}
        \lp\E_{\zeta}\|Q_{0}(x^{(t)}-x^{*})\|_2^{2} - 
           \E_{\zeta}\|Q_{0}(x^{(t+1)}-x^{*})\|_2^{2}
        \rp + \eta G_0^2(1+k\sigma^2) \nonumber\\ 
     &~~~~ +\eta \alpha^2\E_{\zeta}\| Q_0(x^{(t)}-x^{(0)})\|_2^2 - 
                \frac{\alpha}{2}\E_{\zeta}\|Q_{0}(x^{(t)}-x^{(0)})\|_2^{2} + 
                \frac{\alpha}{2}\E_{\zeta}\|Q_{0}(x^{(0)}-x^{*})\|_2^{2}  \nonumber \\
     \leq & ~ \frac{1}{2\eta} \lp\E_{\zeta}\|Q_{0}(x^{(t)}-x^{*})\|_2^{2} - 
                \E_{\zeta}\|Q_{0}(x^{(t+1)}-x^{*})\|_2^{2}  \rp  \\
                &~~~~+\eta G_0^2(1+k\sigma^2)+\frac{\alpha}{2}\E_{\zeta}\|Q_0(x^{(0)}-x^*)\|_2^2 \label{eq:term1},
\end{align}
where we used $\eta\leq\frac{1}{2\alpha}$ at the end. 
\medskip

{\bf  Bounding $(Q_s\nabla f_j(x^{(t)}))^\top (x^{(t)}-x^*)$ :}
We bound the objective above for each $s$ separately. 
By taking expectation over the random $f_j$, we have
\begin{align}
\E_{f_j}(Q_{s}\nabla f_{j}(x^{(t)}))^{\top}(x^{(t)}-x^{*}) = &~ (Q_{s}\nabla F(x^{(t)}))^{\top}(x^{(t)}-x^{*}) \nonumber\\
\leq & ~ \|Q_{s}\nabla F(x^{(t)})\|_2
            \cdot \|Q_s(x^{(t)}-x^{*})\|_2 \nonumber \\
\leq &~ \frac{1}{\alpha_{s}}\|Q_{s}\nabla F(x^{(t)})\|_2^{2} + 
            \frac{\alpha_{s}}{4}\|Q_{s}(x^{(t)}-x^{*})\|_2^{2} \nonumber \\
\leq &~ \frac{G_{2^{s-1}k}^{2}}{\alpha_{s}} + 
            \frac{\alpha_{s}}{2}\|Q_{s}(x^{(t)}-x^{(0)})\|_2^{2} +
            \frac{\alpha_{s}}{2}\|Q_{s}(x^{(0)}-x^{*})\|_2^{2} \label{eq:term2},
\end{align}
where we chose $\alpha_{s}=\alpha s^{-2} 2^{-s}$ and used the bound \eqref{eq:Qbound} and Young's inequality at the end.

\medskip

{\bf Bounding Equation \eqref{eq:mirror_two_term}:} Combining both the terms \eqref{eq:term1} and \eqref{eq:term2} and taking expectation
over all randomness, we have
\begin{align*}
\E [F(x^{(t)})- F(x^{*})] \leq &~~ \frac{1}{2\eta}(
    \E\|Q_{0}(x^{(t)}-x^{*})\|_2^{2} - 
    \E\|Q_{0}(x^{(t+1)}-x^{*})\|_2^{2}) +
    \eta G_{0}^{2}(1+k\sigma^{2})+\frac{\alpha}{2}\E\|Q_{0}(x^{(0)}-x^{*})\|_2^{2}\\
 & +\sum_{s=1}^{S}\frac{G_{2^{s-1}k}^{2}}{\alpha_{s}} + 
    \frac{1}{2}\sum_{s=1}^{S}\alpha_{s}\E\|Q_{s}(x^{(t)}-x^{(0)})\|_2^{2} + 
    \frac{1}{2}\sum_{s=1}^{S}\alpha_s\E\|Q_{s}(x^{(0)}-x^{*})\|_2^{2}\\
\leq &~ \frac{1}{2\eta}(
    \E\|Q_{0}(x^{(t)}-x^{*})\|_2^{2} - 
    \E\|Q_{0}(x^{(t+1)}-x^{*})\|_2^{2}) +
    \eta G_{0}^{2}(1+k\sigma^{2}) + 
    \frac{\alpha}{2}\|x^{(0)}-x^{*}\|_2^{2}\\
 & +\sum_{s=1}^S\frac{G_{2^{s-1}k}^{2}}{\alpha_{s}} + 
    \frac{1}{2}\sum_{s=1}^S\alpha_{s}\E\|Q_{s}(x^{(t)}-x^{(0)})\|_2^{2}.
\end{align*}

Recall $\alpha\cdot\eta\leq 1/2$. Under the other assumptions, by Lemma \ref{lem:radius}, one can show
\begin{align*}
\E\|Q_{s}(x^{(t)}-x^{(0)})\|^{2} & \leq\frac{4 G_{2^{s-1}k}^{2}}{\alpha^{2}}+\frac{2\eta G_{0}^{2}}{\alpha}(1+2^{s}k\sigma^{2})\\
 & \leq\frac{4 G_{2^{s-1}k}^{2}}{\alpha_{s}^{2}}+\frac{2\eta G_{0}^{2}}{\alpha_{s} s^2}(1+k\sigma^2).
\end{align*}
Using $\sum_{s=1}^{\infty} s^{-2} \leq 2$, we have
\begin{align*}
\E F(x^{(t)})-\E F(x^{*}) \leq &~ \frac{1}{2\eta}(
    \E\|Q_{0}(x^{(t)}-x^{*})\|_2^{2} - 
    \E\|Q_{0}(x^{(t+1)}-x^{*})\|_2^{2}
) + \frac{\alpha}{2}\|x^{(0)}-x^{*}\|^{2}\\
&+ \eta G_0^2(1+k\sigma^2)+3\sum_{s=1}^S\frac{G_{2^{s-1}k}^2}{\alpha_s}+2 \eta G_0^2(1+k\sigma^2)\\
\leq  &~ \frac{1}{2\eta}(
    \E\|Q_{0}(x^{(t)}-x^{*})\|_2^{2} -
    \E\|Q_{0}(x^{(t+1)}-x^{*})\|_2^{2}
) + \frac{\alpha}{2}\|x^{(0)}-x^{*}\|_2^{2}\\
&~ +3\eta G_{0}^{2}(1+k\sigma^{2}) +\frac{3}{\alpha}\sum_{s=1}^S s^2 2^{s}G_{2^{s-1}k}^{2}.
\end{align*}
Summing up over $t=1,2,\cdots,T$, by the assumption that $\|x^{(0)}-x^*\|_2\leq D$ and convexity of the function, we have
\begin{align}
\label{eq:utility_eta_alpha}
\E [F(\ox)- F(x^{*})]\leq\frac{D^{2}}{2\eta T}+3\eta G_{0}^{2}(1+k\sigma^{2})+\frac{\alpha}{2}D^{2}+\frac{3}{\alpha}\sum_{s=1}^S s^2 2^{s}G_{2^{s-1}k}^{2}.    
\end{align}

Set the parameters $T=c_1 (n^2 + d \log^2 d)$, $\sigma=\frac{c_2\sqrt{T\log(1/\delta)}}{ n \epsilon}$,
$\eta=\sqrt{\frac{D^{2}}{T\cdot G_{0}^{2}\cdot k\sigma^{2}}}$
and $\alpha=\frac{1}{D}\sqrt{\sum_{s=1}^{S}s^2 2^{s}G_{2^{s-1}k}^{2}}$ for some large constants $c_1,c_2$. 
Note that this choice of parameters satisfies
\begin{align*}
\eta\cdot\alpha
& =
    \sqrt{\frac{D^{2}}{T\cdot G_{0}^{2}\cdot k\sigma^{2}}}\cdot\frac{\sqrt{\sum_{s=1}^{S}s^2 2^{s}G_{2^{s-1}k}^{2}}}{D}\\
 & \leq
    \sqrt{\frac{ G_{0}^{2} (2d) \log^3 (2d)}{T\cdot G_{0}^{2}\cdot k\sigma^{2}}}
    = 
        \frac{n\epsilon}{c_{2}T}\sqrt{\frac{ (2d) \log^3 (2d) }{k \cdot\log(1/\delta)}}\\
 & \leq\frac{n\epsilon\sqrt{ (2d) \log^3 (2d) }}{c_{2}T}\leq\frac{1}{2},
\end{align*}
where we used the fact that $G_k \leq G_0$, $s \le S \le \log (2d)$ , $T \geq n^2 + d \log^2 d$, and $c_2$ is large enough.

Using the parameters we pick, we have
\[
\E [F(\overline{x})- F(x^{*})]\lesssim \frac{G_{0}D\sqrt{k\log(1/\delta)}}{\epsilon n}+D \sqrt{\sum_{s=1}^{S} s^2 2^{s}G_{2^{s-1}k}^{2}}
\]

Moreover,
assuming $G_{k}\leq G_0 k^{-c}$ for some $c>1/2$, we have $\sqrt{\sum_s s^2 2^sG_{2^{s-1}k}^2}\lesssim G_0 / k^{c}$. 
Hence,
\begin{align*}
    \E [F(\overline{x})- F(x^{*})]\lesssim\frac{G_{0}D\sqrt{k\log(1/\delta)}}{\epsilon n}+\frac{G_0D}{k^c}.
\end{align*}

Since the above bound holds for all $k \in \{1, \dots, d\}$, we may optimize it with respect to $k$.
Recall by assumption that $ n\geq \epsilon^{-1} \sqrt{\log(1/\delta)}$. 
Letting
\eqn{
    k = \min \left\{ 
        d,  
        \left\lceil \left( \frac{\epsilon n}{\sqrt{\log(1/\delta)}} \right)^{\frac{2}{1+2c}} \right\rceil
    \right\}
}
yields the bound
\eqn{
    \E[F(\ox;\cD)-F(x^*;\cD)]\lesssim G_{0}D \cdot 
        \bracks{
            \frac{\sqrt{\log(1/\delta)}}{\epsilon n}
        }^{2c/(1+2c)}.
}
\end{proof}

Combining the privacy guarantee in Lemma~\ref{lm:privacy_guarantee} and Lemma~\ref{lm:utility_DPSGD_appendix} directly results in Theorem~\ref{thm:DPERM}.

\subsection{Proof of Theorem~\ref{thm:DPSCO}} \label{app:dpsco_proof}

We study the generalization error of DP-SGD and make use of its stability.
The bound on the excess population loss follows from combining bounds on the excess empirical loss and the generalization error. 
Before stating the proof, we first recall two results in the literature.

\begin{lemma}[{\cite[Lemma 7]{BE02}}]
\label{lm:stab_generalization_error}
Given a learning algorithm $\cA$, a dataset $\cD=\{s_1,\cdots,s_n\}$ formed by $n$ i.i.d. samples drawn from the underlying distribution $\cP$, and we replace one random sample in $\cD$ with a freshly sampled $s'\sim\cP$ to obtain a new neighboring dataset $\cD'$.
One has
\begin{align*}
    \E_{\cD,\cA}[F(\cA(\cD); \cP)-F(\cA(\cD);\cD)]=\E_{\cD,s'\sim\cP,\cA}[f(\cA(\cD);s')-f(\cA(\cD');s')],
\end{align*}
where $\cA(\cD)$ is the output of $\cA$ with input $\cD$.
\end{lemma}

\begin{lemma}[{\cite[Theorem 3.3]{BFGT20}}]
\label{lm:stab_DPSGD}
Suppose Assumption~\ref{assm:G_k} holds, running DP-SGD with step size $\eta$ 
on any two neighboring datasets $\cD$ and $\cD'$ for $T$ steps yields the following bound
\begin{align*}
    \E \sbracks{ \|\ox-\ox'\|_2 } \le 4 G_0 \eta\left(\frac{T}{n}+\sqrt{T}\right),
\end{align*}
where $\ox$ and $\ox'$ are the outputs of DP-SGD with datasets $\cD$ and $\cD'$, respectively.
\end{lemma}

\begin{proof}[Proof of Theorem \ref{thm:DPSCO}]
Let $\ox$ and $\ox'$ be the outputs of DP-SGD when applied to the datasets $\cD$ and $\cD'$, respectively.
$\cD'$ is a neighbor of $\cD$ with one example replaced by $s' \sim \cP$ that is independently sampled. 
Combining Lemma~\ref{lm:stab_generalization_error} and Lemma~\ref{lm:stab_DPSGD} yields
\begin{align*}
    \E [F(\ox; \cP)-F(\ox;\cD)]= &~ \E[f(\ox;s')-f(\ox';s')]\\
    \leq & ~\E [G_0 \|\ox-\ox'\|_2]\\
    \leq & ~ 4G_0^2\eta \bracks{ \frac{T}{n}+\sqrt{T}}.
\end{align*}

Similar to the DP-ERM case, 
by setting $T=c_1 (n^2 + d \log^2 d)$,
$\sigma=\frac{c_2\sqrt{T\log(1/\delta)}}{n\epsilon}$, 
$\eta=\sqrt{\frac{D^{2}}{T \cdot G_{0}^{2}(T/n+k\sigma^{2})}}$
and $\alpha=\frac{1}{D} \sqrt{\sum_{s=1}^S s^2 2^{s}G_{2^{s-1}k}^{2}}$ for some large positive constants $c_1$ and $c_2$, we conclude that $\eta\cdot\alpha\leq 1/2$. Hence, Equation~\eqref{eq:utility_eta_alpha} shows that, for any fixed dataset $\cD$ and any $x^*$ such that $\normtwo{x^{(0)} - x^*} \le D$, we have
\begin{align*}
\E [F(\ox;\cD)- F(x^{*};\cD)]\leq\frac{D^{2}}{2\eta T}+3\eta G_{0}^{2}(1+k\sigma^{2})+\frac{\alpha}{2}D^{2}+\frac{3}{\alpha}\sum_{s=1}^S s^2 2^{s}G_{2^{s-1}k}^{2}.    
\end{align*}

We can rewrite the population loss as follows
\begin{align*}
\E[F(\ox; \cP)-F(x^*; \cP)] =& ~ \E[F(\ox; \cP)-F(\ox;\cD)]+\E_{\cD}[F(\ox;\cD)-F(x^*;\cD)]\\
\leq &~ 
    4G_0^2\eta \bracks{ \frac{T}{n}+\sqrt{T} } + 
    \frac{D^{2}}{2\eta T} + 
    3\eta G_{0}^{2}(1+k\sigma^{2}) + 
    \frac{\alpha}{2}D^{2} + 
    \frac{3}{\alpha}\sum_{s=1}^S s^2 2^{s}G_{2^{s-1}k}^{2}.
\end{align*}

Substituting in the values for parameters $T$, $\sigma$, $\eta$, and $\alpha$ yields
\begin{align*}
    \E[F(\ox; \cP)-F(x^*; \cP)]\lesssim \frac{G_0D}{\sqrt{n}}+\frac{G_{0}D\sqrt{k\log(1/\delta)}}{\epsilon n}+D\sqrt{\sum_{s=1}^{S}s^2 2^{s}G_{2^{s-1}k}^{2}} 
\end{align*}
for all $k \in [d]$.

Similarly, if we have $G_k\leq G_0 k^{-c}$ for some $c>1/2$, and in addition $n > \epsilon^{-1} \log(1/\delta)$, it immediately follows that
\begin{align*}
    \E[F(\ox; \cP)-\min_x F(x; \cP)]\lesssim \frac{G_0D}{\sqrt{n}}+G_{0}D\cdot \bracks { \frac{\sqrt{\log(1/\delta)}}{\epsilon n} }^{2c/(1+2c)}.
\end{align*}
This completes the proof.
\end{proof}

\section{Numerical Experiments}\label{sec:experiments}
The aim of this section is twofold.
In Section~\ref{sec:synthetic}, we study a synthetic example that matches our theoretical assumptions and show that DP-SGD attains dimension-independent empirical and population loss when the sequence of restricted Lipschitz coefficients decays rapidly---even when gradients span the entire ambient space.
In Section~\ref{sec:lm}, we study a stylized example of privately fine-tuning large language models. 
Building on the previous theory, we provide insights as to why dense fine-tuning can yield good performance. 

\subsection{Synthetic Example: Estimating the Generalized Geometric Median}\label{sec:synthetic}
We privately estimate the geometric median which minimizes the average Mahalanobis distance. 
Specifically, let $x_i \in \R^d$ for $i \in [n]$ be feature vectors drawn i.i.d. from some distribution $P_x$, each of which is treated as an individual record. 
Denote the entire dataset as $\mathcal{D} = \{ x_i\}_{i=1}^n$.
Subject to differential privacy, we perform the following optimization
\eq{\label{eq:synth}
\min_{ x \in \R^d} F_\alpha (x)
	=
		\frac{1}{n} \sum_{i=1}^n f_i (x) + \frac{\alpha}{2} \normtwo{ x - x^{(0)} }^2
	= 
		\frac{1}{n} \sum_{i=1}^n \left\| x - x_i \right\|_A  + \frac{\alpha}{2} \normtwo{ x - x^{(0)} }^2,
}
where we adopt the shorthand $f_i(x) = f(x; x_i) = \norm{x - x_i}_A$.
When $A = I_d$ and $\alpha=0$ (without the regularization term), the problem reduces to estimating the usual geometric median (commonly known as center of mass). 

For this example, individual gradients are bounded since $\| \nabla f_i( x ) \|_2 = \| A (x - x_i)  / \| x - x_i \|_A \|_2 \le \lambda_1(A^{1/2}) = G_0$.
More generally, the restricted Lipschitz coefficients of $F(x)$ are the eigenvalues of $A^{1/2}$, since 
\eqn{
	\| Q_k \nabla F(x) \|_2
=
	 \left\|Q_k A^{1/2} \frac{1}{n} \sum_{i=1}^n \frac{ A^{1/2}( x - x_i ) }{ \| x - x_i \|_A } \right\|_2
\le
	\| Q_k A^{1/2} \|_\text{op} = \lambda_{k+1}(A^{1/2}) = G_k, 
}
where $Q_k = I - P_k$ is chosen to be the rank $(d-k)$ orthogonal projection matrix that projects onto the subspace spanned by the bottom $(d-k)$ eigenvectors of $A^{1/2}$.

To verify our theory, we study the optimization and generalization performance of DP-SGD for minimizing~\eqref{eq:synth} under Mahalanobis distances induced by different $A$ as the problem dimension grows. 
The optimization performance is measured by the final training error, and the generalization performance is measured by the population quantity $\mathbb{E}_{x \sim P_x, \ox} [ \| \ox - x \|_A ]$, where $\ox$ denotes the random output of DP-SGD.
We study the dimension scaling behavior for $A$ being one of 
\eqn{
A_{\text{const}} = \text{diag}( 1, \dots, 1 ), \;\;
A_{\text{sqrt}} = \text{diag}( 1, 1 / \sqrt{2}, \dots, 1 / \sqrt{d} ), \;\; 
A_{\text{linear}} = \text{diag} ( 1, 1 / 2, \dots, 1 / d ),
}
where $\text{diag}: \R^d \to \R^{d \times d}$ maps vectors onto square matrices with inputs on the diagonal.
In all cases, the span of gradients $\text{span}( \{ \nabla F(x) \} )$ is the ambient space $\R^d$, since $A$ is of full rank.
To ensure the distance from the initial iterate $\beta^{(0)} = 0$ to the optimum is the same for problem instances of different dimensions, we let feature vectors $\{x_i\}_{i=1}^n$ take zero values in any dimension $k > d_{\text{min}}$, where $d_{\text{min}}$ is the dimension of the smallest problem in our experiments.
Our theoretical bounds suggest that when the sequence of restricted Lipschitz coefficients is constant (when $A = A_{\text{const}}$), the excess empirical loss grows with the problem dimension, whereas when the sequence of $k$th-Lipschitz constants rapidly decays with $k$ (when $A = A_{\text{sqrt}}$ or $A = A_{\text{linear}}$), the excess empirical loss does not grow beyond a certain problem dimension. 
Figure~\ref{fig:toy} empirically captures this phenomenon. 
We include additional experimental setup details in Appendix~\ref{app:exp_synth}.
\begin{figure}[t]
\begin{center}
\begin{minipage}[t]{0.48\linewidth}
\centering
{\includegraphics[width=0.98\textwidth]{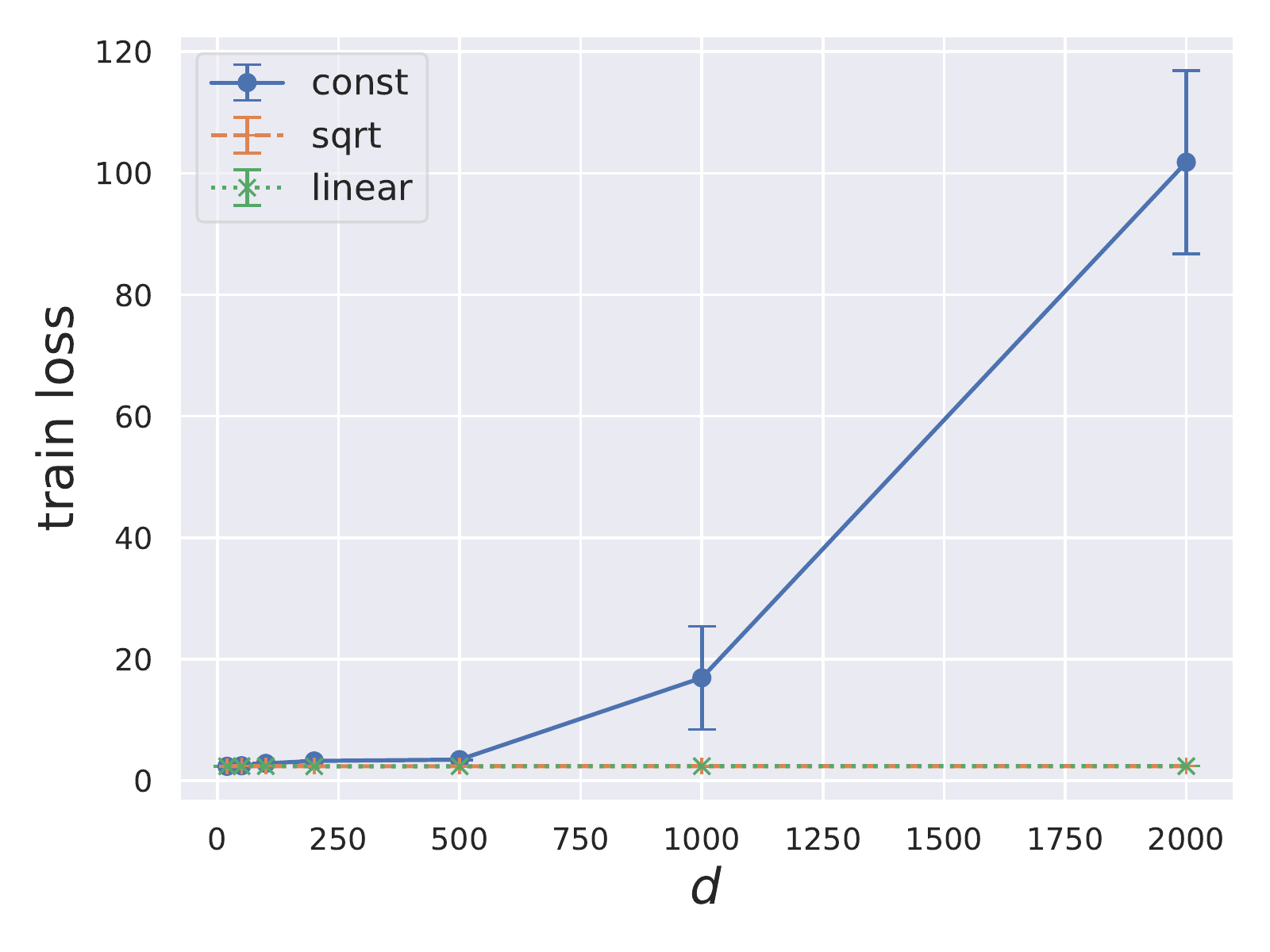}}
(a) empirical loss
\end{minipage}
\begin{minipage}[t]{0.48\linewidth}
\centering
{\includegraphics[width=0.98\textwidth]{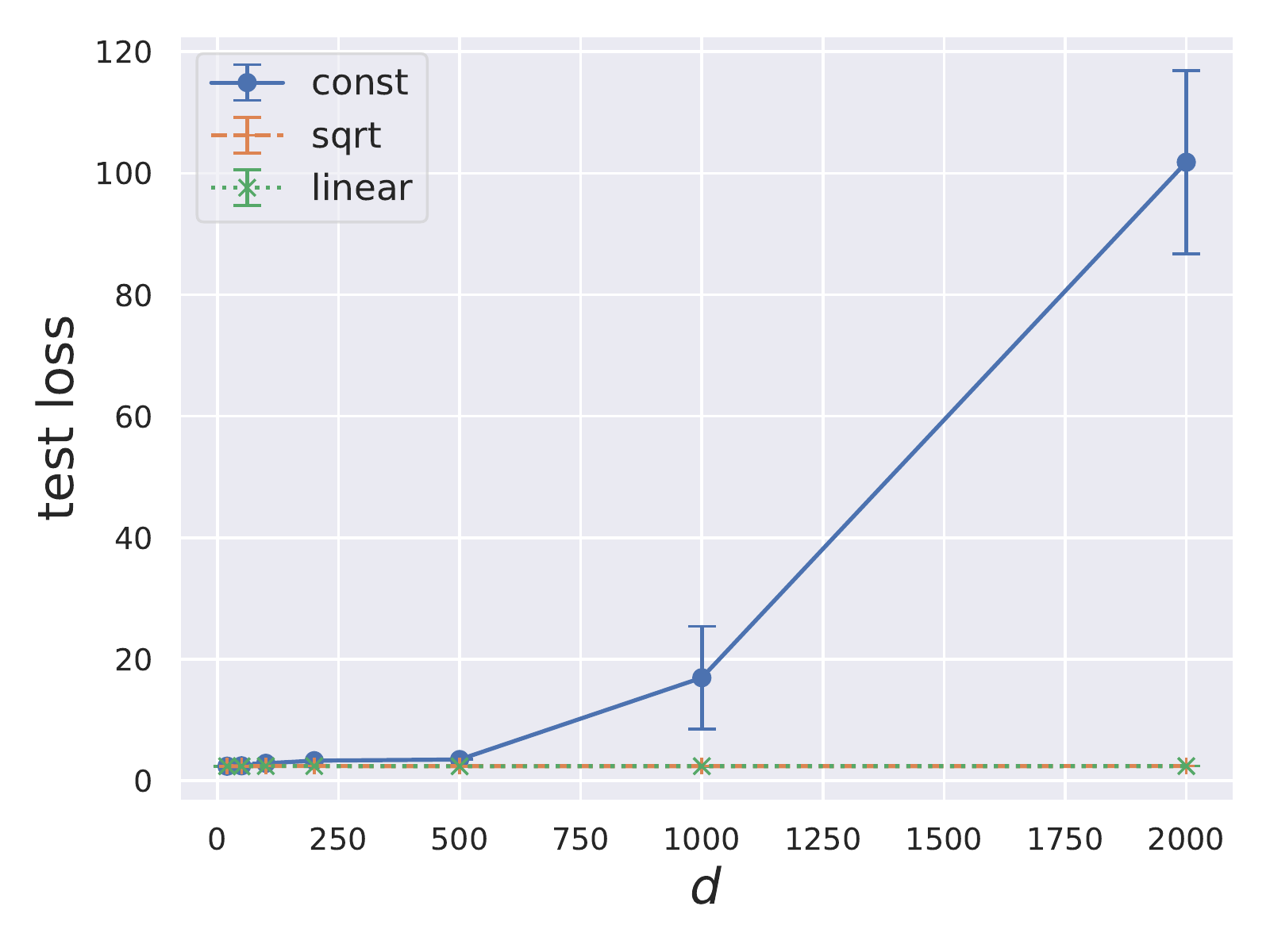}}
(b) (estimated) population loss
\end{minipage}
\end{center}
\caption{
The empirical and population losses grow with increasing problem dimension when the sequence of restricted Lipschitz coefficients remain constant. On the other hand, these losses remain almost constant when the sequence of restricted Lipschitz coefficients decays rapidly.
Error bars represent one standard deviation over five runs of DP-SGD with the same hyperparameters which were tuned on separate validation data.
For the same $A$, the optimal training error $\min_{x\in\R^d} F(x)$ is the same for problem instances with different dimensions (thus errors do not scale if learning was non-private).
Each training run was performed with $\epsilon=2$, $\delta=10^{-6}$, and $n=10000$. 
}
\label{fig:toy}
\end{figure}

\subsection{Why Does Dense Fine-Tuning Work Well for Pretrained Language Models?}\label{sec:lm}
Stated informally, our bounds in Theorem~\ref{thm:DPSCO} imply that DP-SGD obtains dimension-independent errors if gradients approximately reside in a subspace much smaller than the ambient space.
Inspired by these results for the convex case, we now turn to study dense language model fine-tuning~\cite{li2021large} and provide a possible explanation for their recent intriguing success --- fine-tuning gigantic parameter vectors frequently results in moderate performance drops compared to non-private learning.

In the following, we present evidence that gradients obtained through fine-tuning mostly lie in a small subspace.
We design subsequent experiments to work under a simplified setup. 
Specifically, we fine-tune DistilRoBERTa~\cite{sanh2019distilbert,liu2019roberta} under $\epsilon=8$ and $\delta = \nicefrac{1}{n^{1.1}}$ for sentiment classification on the SST-2 dataset~\cite{socher2013recursive}.
We reformulate the label prediction problem as templated text prediction~\cite{li2021large}, and fine-tune only the query and value matrices in attention layers.

We focus on fine-tuning these specific parameter matrices due to the success of LoRA for non-private learning~\cite{hu2021lora} which focuses on adapting the attention layers. Unlike LoRA, we fine-tune all parameters in these matrices rather than focusing on low-rank updates.
This gives a setup that is lightweight enough to run spectral analyses computationally tractably but retains enough parameters ($\approx7$ million) such that a problem of similar scale outside of fine-tuning results in substantial losses in utility.\footnote{For instance, an off-the-shelf ResNet image classifier has 10 to 20+ million parameters. A plethora of works report large performance drops when training these models from scratch~\cite{YZCL21,luo2021scalable,de2022unlocking}.}
For our setup, DP-SGD obtains a dev set accuracy approximately of $90\%$ and $92\%$, privately and non-privately, respectively.
These numbers are similar to previous results obtained with the same pretrained model~\cite{yu2021differentially,li2021large}. We include the full experimental protocol and additional results in Appendix~\ref{app:exp_privlm}. 

\begin{figure}[thb]
\begin{center}
\begin{minipage}[t]{0.48\linewidth}
\centering
{\includegraphics[width=0.98\textwidth]{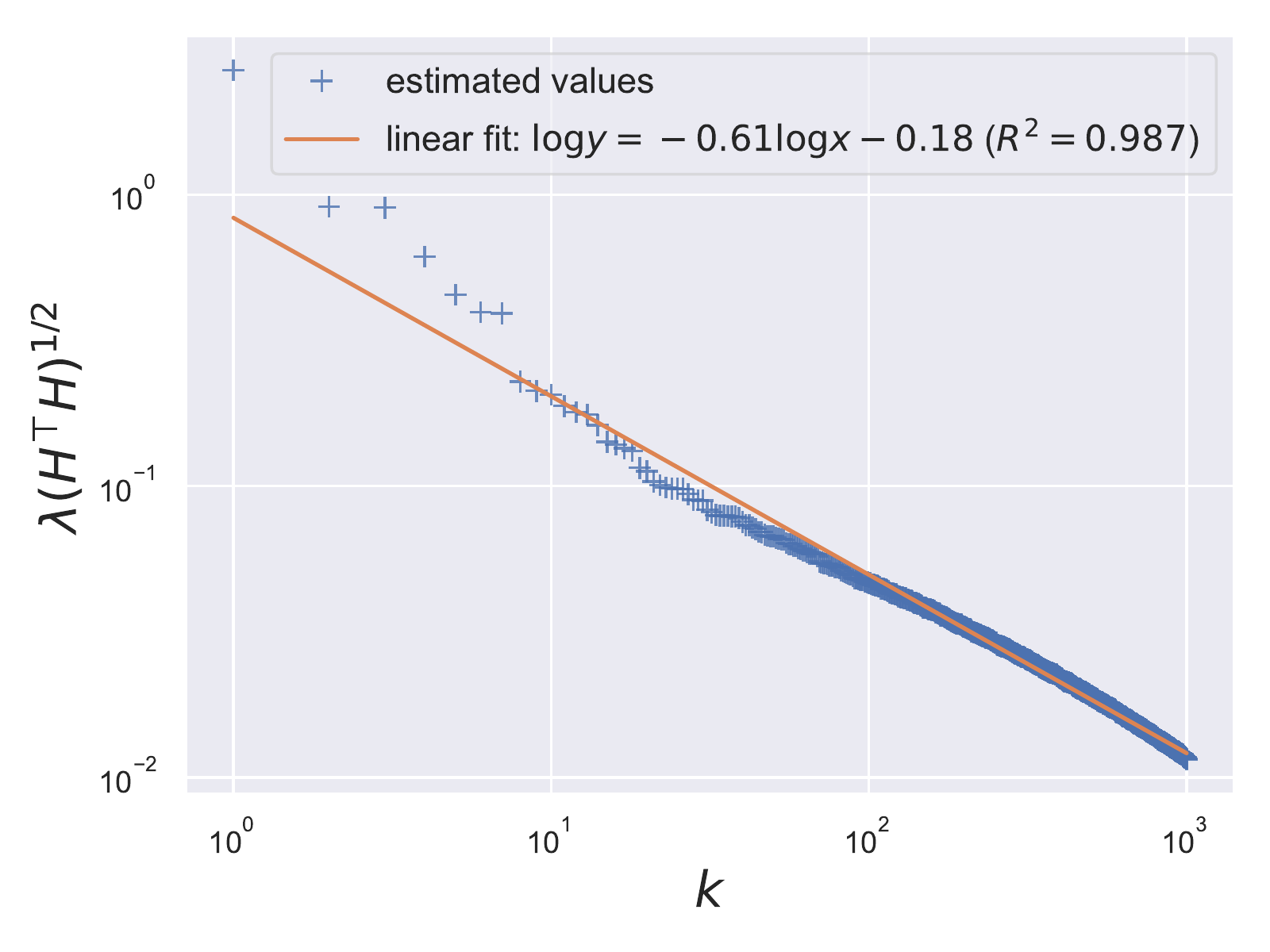}}
\footnotesize{(a) singular values decay with rank}
\end{minipage}
\begin{minipage}[t]{0.48\linewidth}
\centering
{\includegraphics[width=0.98\textwidth]{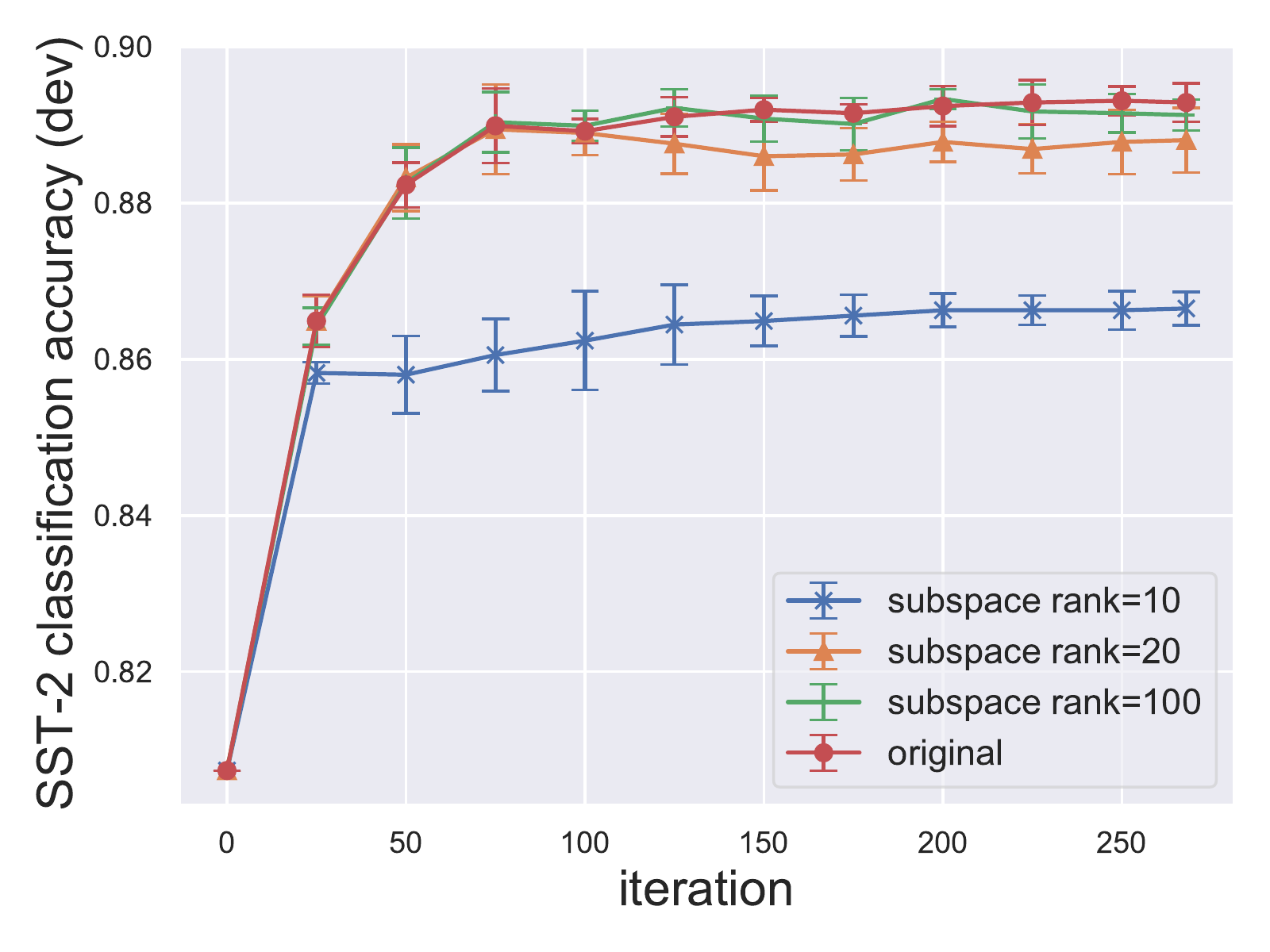}}
\footnotesize{(b) retrain in fixed subspace}
\end{minipage}
\end{center}
\caption{
Gradients obtained through fine-tuning are controlled by a few principal components.
\emph{Left:} Singular values decay rapidly with their rank.
\emph{Right:} Retraining with gradients projected onto a subspace is sufficient to recover original performance.
}
\label{fig:pca}
\end{figure}

To provide evidence for the small subspace hypothesis, we sample gradients during fine-tuning and study their principal components.
Specifically, we ``over-train'' by privately fine-tuning for $r = 2 \times 10^3$ updates and collect all the non-privatized average clipped gradients along the optimization trajectory.
While fine-tuning for 200 and 2k updates have similar final dev set performance under our hyperparameters, the increased number of steps allows us to collect more gradients around the converged solution. 
This yields a gradient matrix $H \in \R^{r \times p}$, where $p \approx 7 \times 10^6$ is the size of the parameter vector. 
We perform PCA for $H$ with the orthogonal iteration algorithm~\cite{demmel1997applied} and visualize the set of estimated singular values $\sigma_i(H) = \lambda_i(H^\top H)^{1/2}$ in terms of both (i) the density estimate, and (ii) their relation with the rank.
Figure~\ref{fig:pca} (a) shows the top 1000 singular values sorted and plotted against their rank $k$ and the least squares fit on log-transformed inputs and outputs.
The plot displays few large singular values which suggests that gradients are controlled through only a few principal directions.
The linear fit suggests that singular values decay rapidly (at a rate of approximately $k^{-0.6}$).

To study the effects that different principal components have on fine-tuning performance, we further perform the following re-training experiment. 
Given the principal components, we privately re-fine-tune with gradients projected onto the top $k \in \{10, 20, 100\}$ components.
Note that this projection applies only to the (non-privatized) average clipped gradients and the isotropic DP noise is still applied to all dimensions.
Figure~\ref{fig:pca} (b) shows that the original performance can be attained by optimizing within a subspace of only dimension $k=100$, suggesting that most of the dimensions of the 7 million parameter vector encode a limited learning signal.

While these empirical results present encouraging insights for the dimension-independent performance of fine-tuning, we acknowledge that this is not a complete validation of the restricted Lipschitz continuity condition and fast decay of coefficients (even locally near the optimum).
We leave a more thorough analysis with additional model classes and fine-tuning tasks to future work.

\section{Related Work}
DP-ERM and DP-SCO are arguably the most well-studied areas of differential privacy~\cite{chaudhuri2011differentially,kifer2012private,bassily2014private,song2013stochastic,wang17amd,fts17,bassily2019private,mcmahan2017learning,zzmw17,WLKCJN17,FKT20,iyengar2019towards,BFGT20,song2020characterizing,ll21,afkt21,bgn21,gtu22,gll22}. Tight dependence on the number of model parameters and the number of samples is known for both DP-ERM~\cite{bassily2014private} and DP-SCO~\cite{bassily2019private}. 
In particular, for the error on general convex losses, an explicit polynomial dependence on the number of optimization variables is necessary.
However, it is shown that if gradients lie in a fixed low-rank subspace $M$, the dependence on dimension $d$ can be replaced by ${\sf rank}(M)$ which can be significantly smaller~\cite{jain2014near,song2020characterizing}. 
We extend this line of work to show that under a weaker assumption (restricted Lipschitz continuity with decaying coefficients) one can obtain analogous error guarantees that are independent of $d$, but do not require the gradients of the loss to strictly lie in any fixed low-rank subspace $M$. 
As a consequence, our results provide a plausible explanation for the empirical observation that dense fine-tuning can be effective and that fine-tuning a larger model under DP can generally be more advantageous in terms of utility than fine-tuning a smaller model~\cite{li2021large,yu2021differentially}. 
A concurrent work shows that the standard dimension dependence of DP-SGD can be replaced by a dependence on the trace of the Hessian assuming the latter quantity is uniformly bounded~\cite{ma2022dimension}.

A complementary line of work designed variants of DP-SGD that either explicitly or implicitly control the subspace in which gradients are allowed to reside~\cite{PDA-DPMD,liu2021leveraging,asi2021private,KRRT21,YZCL21}. 
They demonstrated improved dependence of the error on the dimension if the true gradients lie in a ``near'' low-rank subspace. 
Our results are incomparable to this line of work because of two reasons: (i) Our algorithm is vanilla DP-SGD and does not track the gradient subspace either explicitly or implicitly, and hence does not change the optimization landscape. 
Our improved dependence on dimensions is an artifact of the analysis. 
(ii) Our analytical results do not need the existence of any public data to obtain tighter dependence on dimensions. All prior works mentioned above need the existence of public data to demonstrate any improvement.

On the empirical front, past works have observed that for image classification tasks, gradients of ConvNets converge to a small subspace spanned by the top directions of the Hessian. In addition, this span remains stable for long periods of time during training~\cite{gur2018gradient}.
While insightful, this line of work does not look at language model fine-tuning.
Another line of work measures for language model fine-tuning the \emph{intrinsic dimension}---the minimum dimension such that optimizing in a randomly sampled subspace of such dimension approximately recovers the original performance~\cite{li2018measuring,aghajanyan2020intrinsic}. 
We note that a small intrinsic dimension likely suggests that gradients are approximately low rank. 
Yet, this statement should not be interpreted as a strict implication, since the notion of intrinsic dimension is at best vaguely defined (e.g., there's no explicit failure probability threshold over the randomly sampled subspace in the original statement), and the definition involves not a fixed subspace but rather a randomly sampled one.

\section{Conclusion}
We made an attempt to reconcile two seemingly conflicting results: (i) in private convex optimization, errors are predicted to scale proportionally with the dimension of the learning problem; while (ii) in empirical works on large-scale private fine-tuning through DP-SGD, privacy-utility trade-offs become better with increasing model size.
We introduced the notion of restricted Lipschitz continuity, with which we gave refined analyses of DP-SGD for DP-ERM and DP-SCO. 
When the magnitudes of gradients projected onto diminishing subspaces decay rapidly, our analysis showed that excess empirical and population losses of DP-SGD are independent of the model dimension. 
Through preliminary experiments, we gave empirical evidence that gradients of large pretrained language models obtained through fine-tuning mostly lie in the subspace spanned by a few principal components. 
Our theoretical and empirical results together give a possible explanation for recent successes in large-scale differentially private fine-tuning.

Given our improved upper bounds on the excess empirical and population risks for differentially private convex learning, it is instructive to ask if such bounds are tight in the mini-max sense.
We leave answering this inquiry to future work.
In addition, while we have presented encouraging empirical evidence that fine-tuning gradients mostly lie in a small subspace, more work is required to study the robustness of this phenomenon with respect to the model class and fine-tuning problem.
Overall, we hope that our work leads to more research on understanding conditions under which DP learning does not degrade with increasing problem size, and more generally, how theory can inform and explain the practical successes of differentially private deep learning.

\subsubsection*{Acknowledgments}
We thank Guodong Zhang for helpful discussions and comments on an early draft.
XL is supported by a Stanford Graduate Fellowship.
Lee and Liu are partially supported by NSF awards CCF-1749609, DMS-1839116, DMS-2023166, CCF-2105772, a Microsoft Research Faculty Fellowship, a Sloan Research Fellowship, and a Packard Fellowship.

\addcontentsline{toc}{section}{References}
\bibliographystyle{alpha}
\bibliography{main}

\newpage
\appendix
\section*{Appendix}

\section{Protocol for Synthetic Example Experiments in Section~\ref{sec:synthetic}} \label{app:exp_synth}
We detail the construction of the synthetic example in Section~\ref{sec:synthetic}.
The training and test sets of this example both have $n_\text{train} = n_\text{test} = 10000$ instances.
Each instance $x_i$ is sampled from a distribution where the first $d_\text{min}=10$ coordinates are all multi-variate normal distributions with mean and standard deviation both being $1$. All remaining coordinates are constantly $0$.
This ensures the optimal non-private training losses for problems of different dimensions are the same.

\section{Protocol and Additional Fine-Tuning Experiments for Section~\ref{sec:lm}} \label{app:exp_privlm}

\subsection{Experimental Protocol}
For experiments in Section~\ref{sec:lm}, we fine-tuned the DistilRoBERTa model with $(8, \nicefrac{1}{n^{1.1}})$-DP on the SST-2 training set with $n \ge 60000$ examples and measured performance on the companion dev set.
We used the exact set of hyperparameters presented in~\cite{li2021large} for this task.
We repeated our re-training experiments over five independent random seeds.
Fine-tuning for 3 epochs on this task takes 10 minutes on an A100-powered machine with sufficient RAM and CPU cores.

Our spectral analysis relies on running the orthogonal iteration algorithm on the set of collected gradients along the private fine-tuning trajectory~\cite{demmel1997applied}.\footnote{By gradients, we always mean the average of clipped per-example gradients --- before adding noise --- in this section.}
Unlike other numerical algorithms for estimating eigenvalues of a matrix, orthogonal iteration provably produces the set of eigenvalues that are largest in absolute value (and corresponding eigenvectors, if the underlying matrix is normal) in the limit.\footnote{The orthogonal iteration algorithm is also known as simultaneous iteration or subspace iteration.}
Recall that we needed the top eigenvectors for projection in our re-training experiment.\footnote{
Note that one commonly used algorithm in the neural net spectral analysis literature --- Lanczos iteration~\cite{ghorbani2019investigation} --- does not guarantee that the top eigenvalues are produced, even though its spectral estimates are frequently deemed accurate~\cite{granziol2019deep}.
}
By default, we run orthogonal iteration for ten iterations. 
We show in the following that our results are insensitivity to the number of iterations used in the orthogonal iteration algorithm.
Each orthogonal iteration takes $10$ minutes on an A100-powered machine with $r=4000$ gradient samples and $k=1000$ principal components for the DistilRoBERTa experiments.

\subsection{Robustness of Results}

\begin{figure}[thb]
\begin{center}
\begin{minipage}[t]{0.32\linewidth}
\centering
{\includegraphics[width=0.98\textwidth]{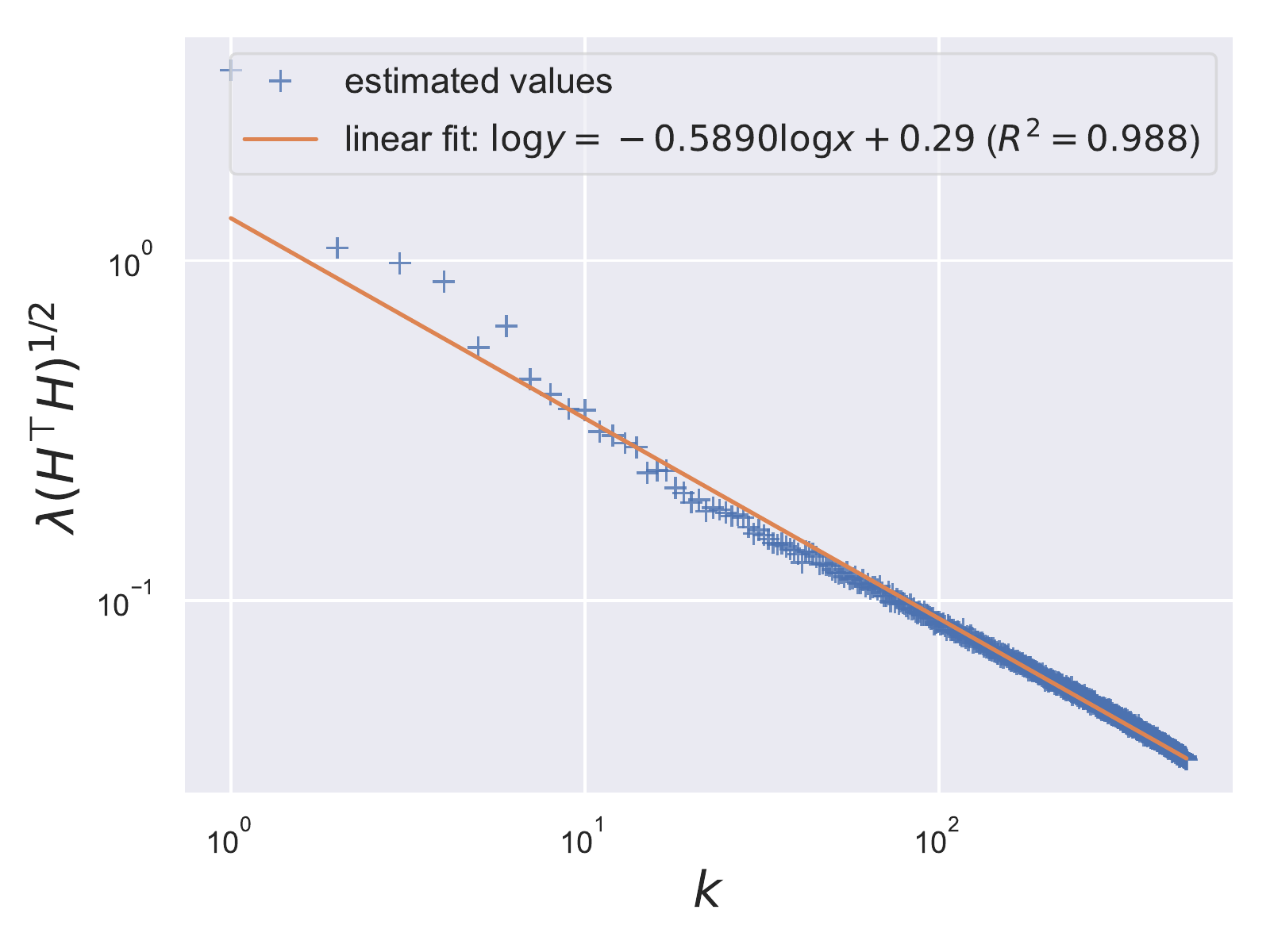}}
(a) $T=10$
\end{minipage}
\begin{minipage}[t]{0.32\linewidth}
\centering
{\includegraphics[width=0.98\textwidth]{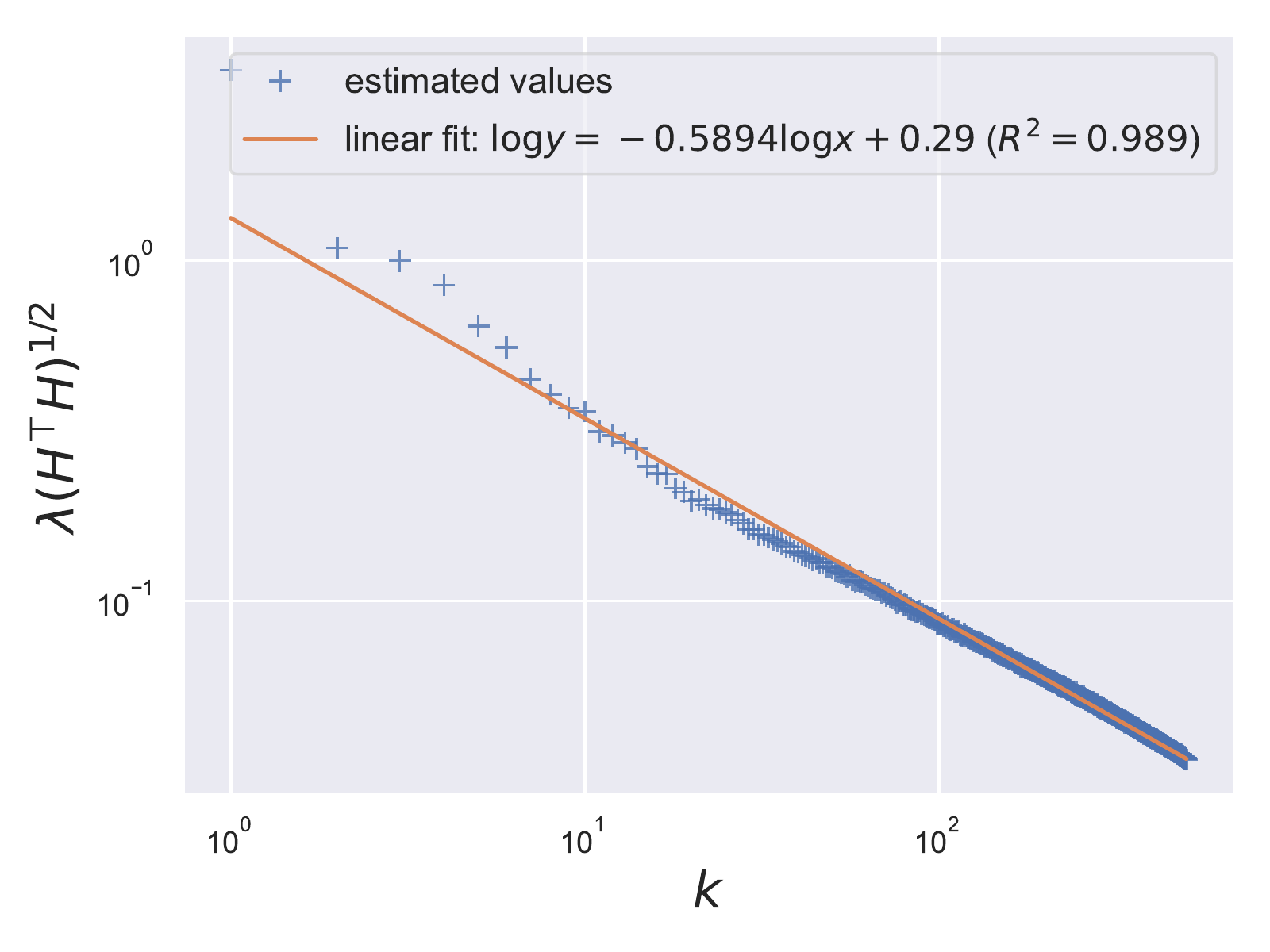}}
(b) $T=50$
\end{minipage}
\begin{minipage}[t]{0.32\linewidth}
\centering
{\includegraphics[width=0.98\textwidth]{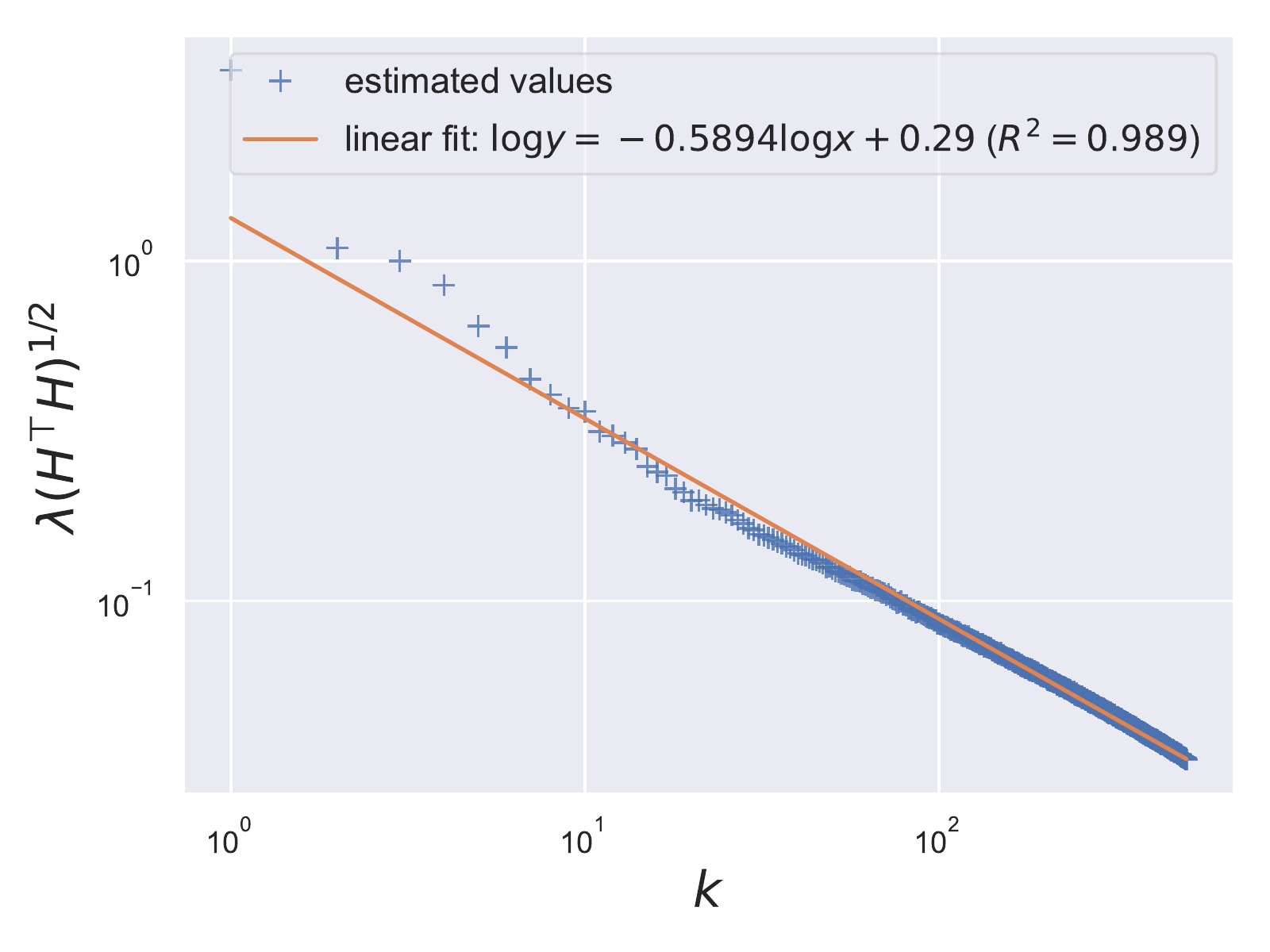}}
(c) $T=100$
\end{minipage}
\end{center}
\caption{
The eigenspectrum remains stable with different number of iterations $T$. 
}
\label{fig:num_power_iter_robust}
\end{figure}

\paragraph{Robustness to the number of orthogonal iterations.}
The orthogonal iteration algorithm is a generalization of the power method that simultaneously produces estimates of multiple eigenvalues.
Its convergence is known to be sensitivity to the gap between successive eigenvalues.
The algorithm converges slowly if consecutive eigenvalues (with the largest absolute values) are close in absolute value.
To confirm that our results aren't sensitivity to the choice of the number of iterations, we visualize the top 500 eigenvalues for the orthogonal iteration algorithm is run for different number of updates.
Figure~\ref{fig:num_power_iter_robust} shows that the linear fit to the top $500$ eigenvalues remains stable across different number of orthogonal iterations $T$.
Notably, $T=10$ produces similar results as $T=100$. 
These results were obtained with $r=4000$ gradients.

\begin{figure}[thb]
\begin{center}
\begin{minipage}[t]{0.45\linewidth}
\centering
{\includegraphics[width=0.98\textwidth]{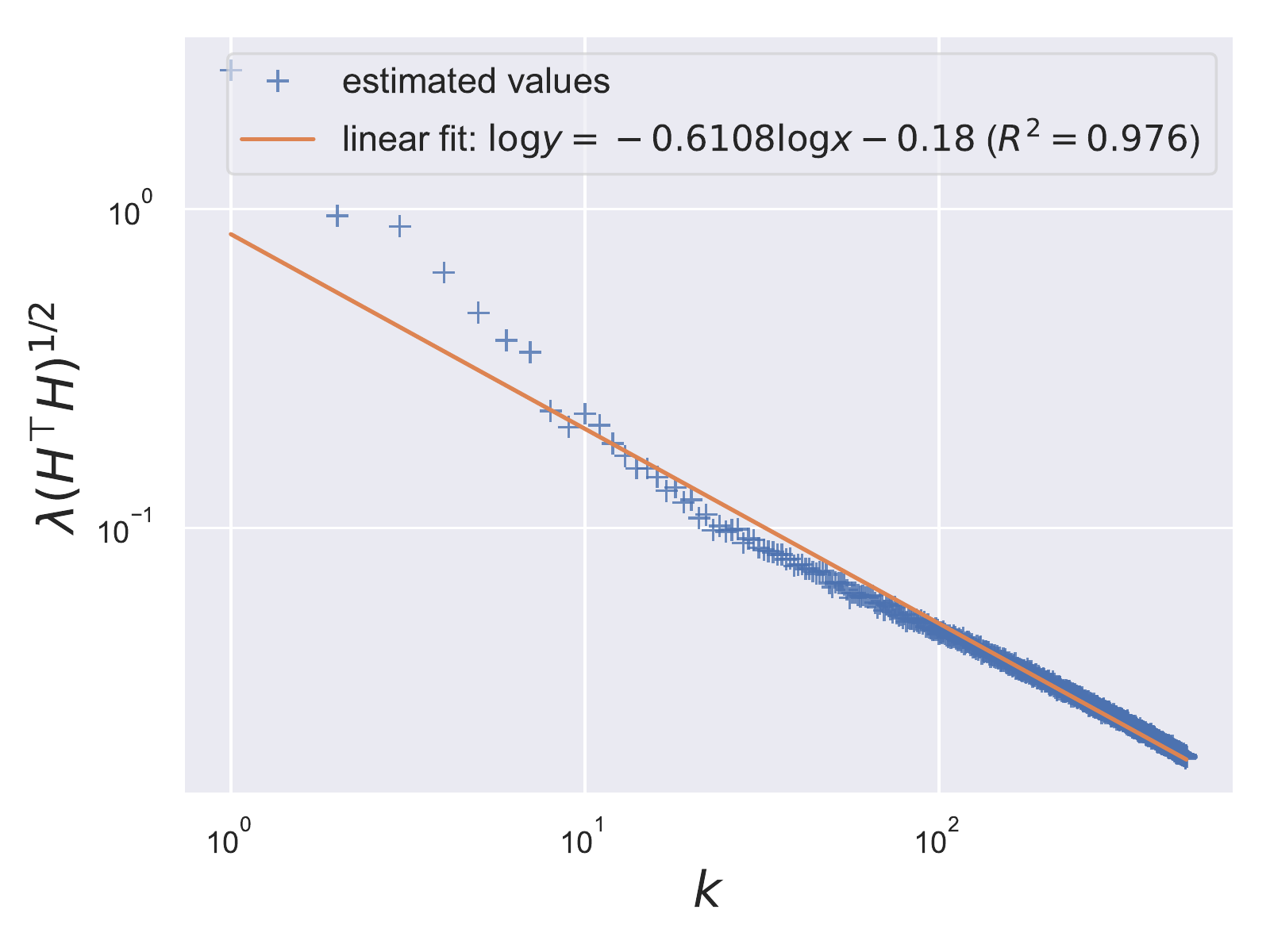}}
(a) $r=1000$
\end{minipage}
\begin{minipage}[t]{0.45\linewidth}
\centering
{\includegraphics[width=0.98\textwidth]{figs/privlm/roberta/npi_000010/eigenvalue-linfit.pdf}}
(b) $r=4000$
\end{minipage}
\end{center}
\caption{
The fast decaying trend of the eigenspectrum remains stable with different number of samples $r$ used for the spectral analysis.
}
\label{fig:num_grad_sample_robust}
\end{figure}

\paragraph{Robustness to the number of gradient samples.}
We further ran experiments with different numbers of gradient samples $r$ collected along the fine-tuning trajectory, and plot the top $500$ eigenvalues.
Figure~\ref{fig:num_grad_sample_robust} shows that while the slope and intercept of the fitted line in log-log space changes, the change is moderate.
Notably, the decaying trend of the top eigenvalues remains stable.

\begin{figure}[thb]
\begin{center}
\begin{minipage}[t]{0.45\linewidth}
\centering
{\includegraphics[width=0.98\textwidth]{figs/privlm/roberta/less_samples_no_trim_front/eigenvalue-linfit.pdf}}
(a) Gradients at iterations $0$ to $1000$
\end{minipage}
\begin{minipage}[t]{0.45\linewidth}
\centering
{\includegraphics[width=0.98\textwidth]{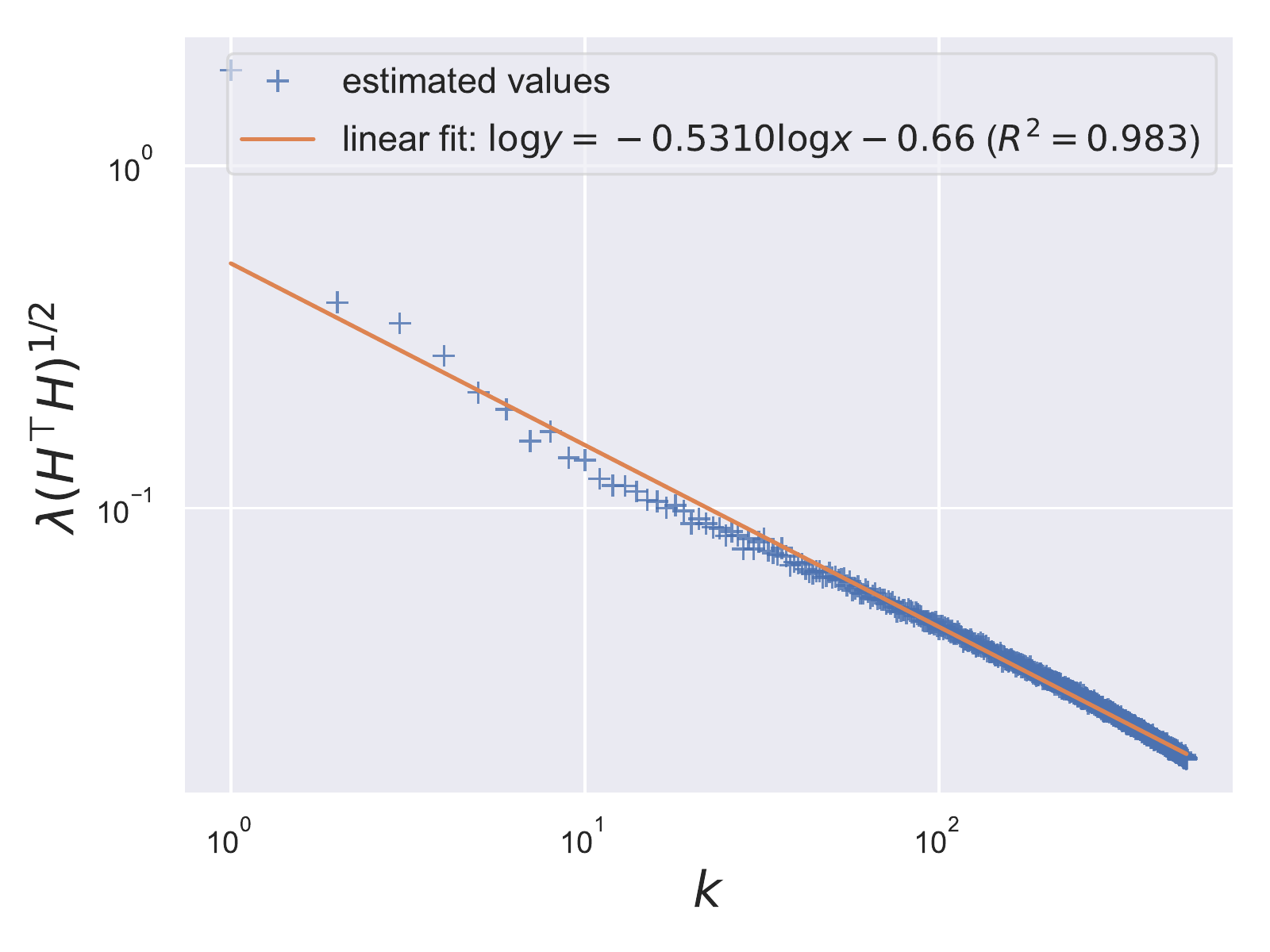}}
(b) Gradients at iterations $300$ to $1300$
\end{minipage}
\end{center}
\caption{
The rapidly decaying trend of the eigenspectrum remains stable with different sampling strategies.
}
\label{fig:trunc_strategy_robust}
\end{figure}

\paragraph{Robustness to the gradient sampling strategy.}
We observe that gradients at the beginning of fine-tuning tend to be larger in magnitude than gradients collected later on along the optimization trajectory.
To eliminate the potential confounder that the top principal components are solely formed by the initial few gradients evaluated during fine-tuning, we re-ran the spectral analysis experiment without the initial gradients.
Specifically, we performed PCA for the gradients evaluated from step 300 to step 1300 during private fine-tuning, and compared the distribution of top eigenvalues returned from this setup to when we used the first 1000 gradients.
Note the dev set accuracy converged to $\approx 90\%$ by step 200. 
Figure~\ref{fig:trunc_strategy_robust} shows that while the slope and intercept of linear fits are slightly different in the new setup compared to the old setup (when all gradients along the fine-tuning trajectory were used for PCA), that the eigenvalues follow a rapidly decaying trend remains true under both setups.

\begin{figure}[thb]
\begin{center}
\begin{minipage}[t]{0.45\linewidth}
\centering
{\includegraphics[width=0.98\textwidth]{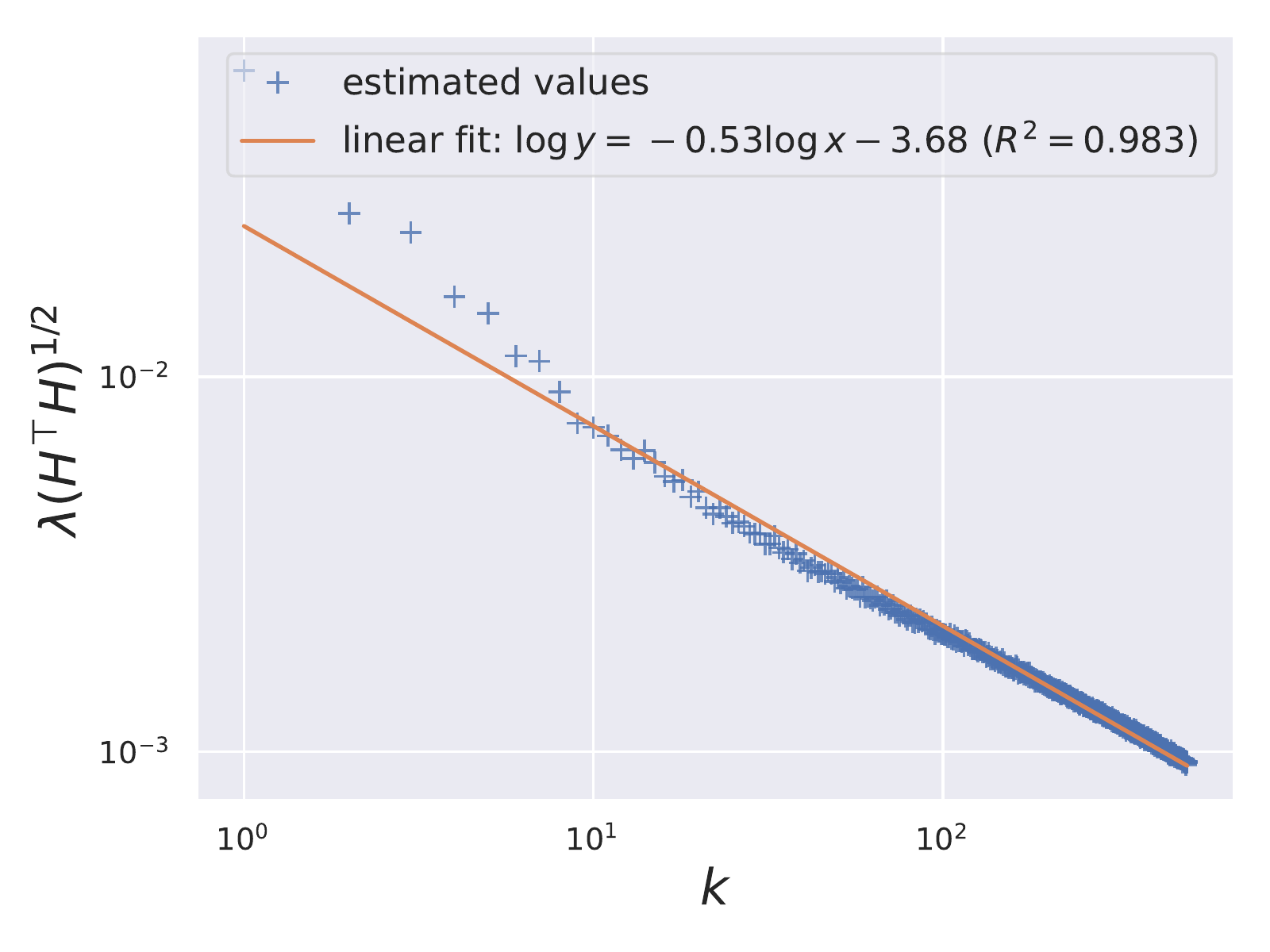}}
\footnotesize{(a) singular values decay with rank}
\end{minipage}
\begin{minipage}[t]{0.45\linewidth}
\centering
{\includegraphics[width=0.98\textwidth]{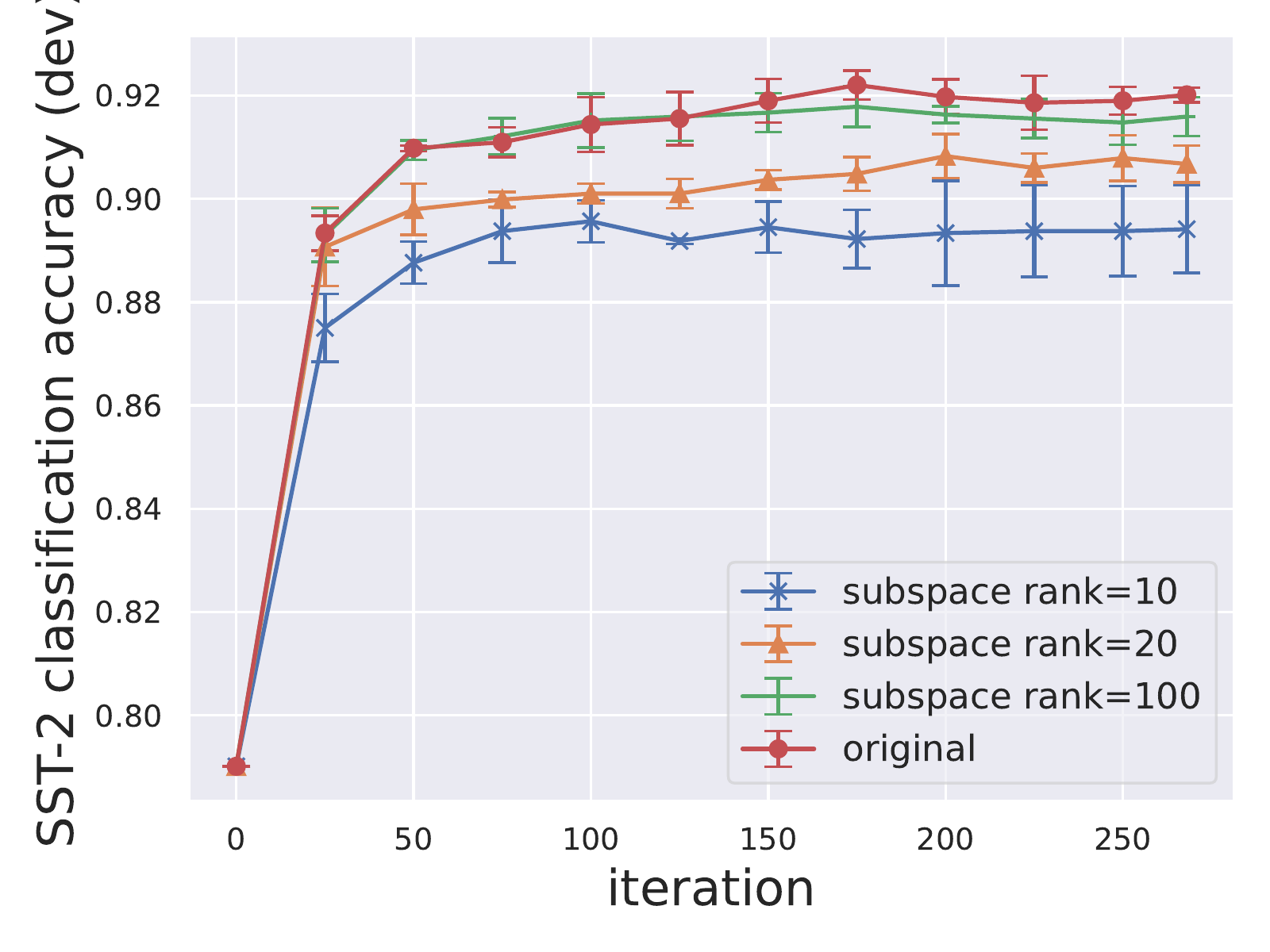}}
\footnotesize{(b) retrain in fixed subspace}
\end{minipage}
\end{center}
\caption{
Experiments for Roberta-base.
}
\label{fig:pca_roberta_base}
\end{figure}

\begin{figure}[thb]
\begin{center}
\begin{minipage}[t]{0.45\linewidth}
\centering
{\includegraphics[width=0.98\textwidth]{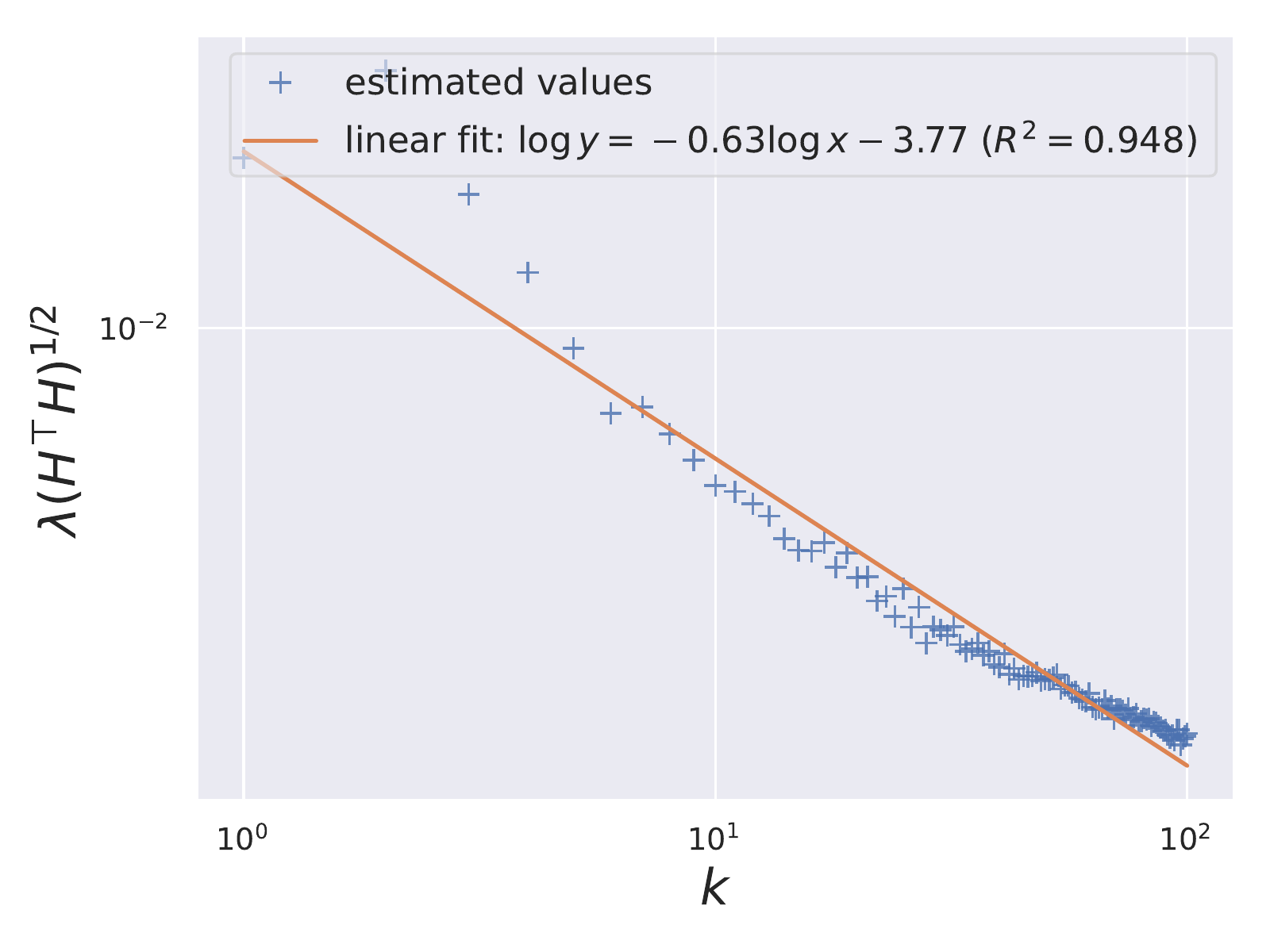}}
\footnotesize{(a) singular values decay with rank}
\end{minipage}
\begin{minipage}[t]{0.45\linewidth}
\centering
{\includegraphics[width=0.98\textwidth]{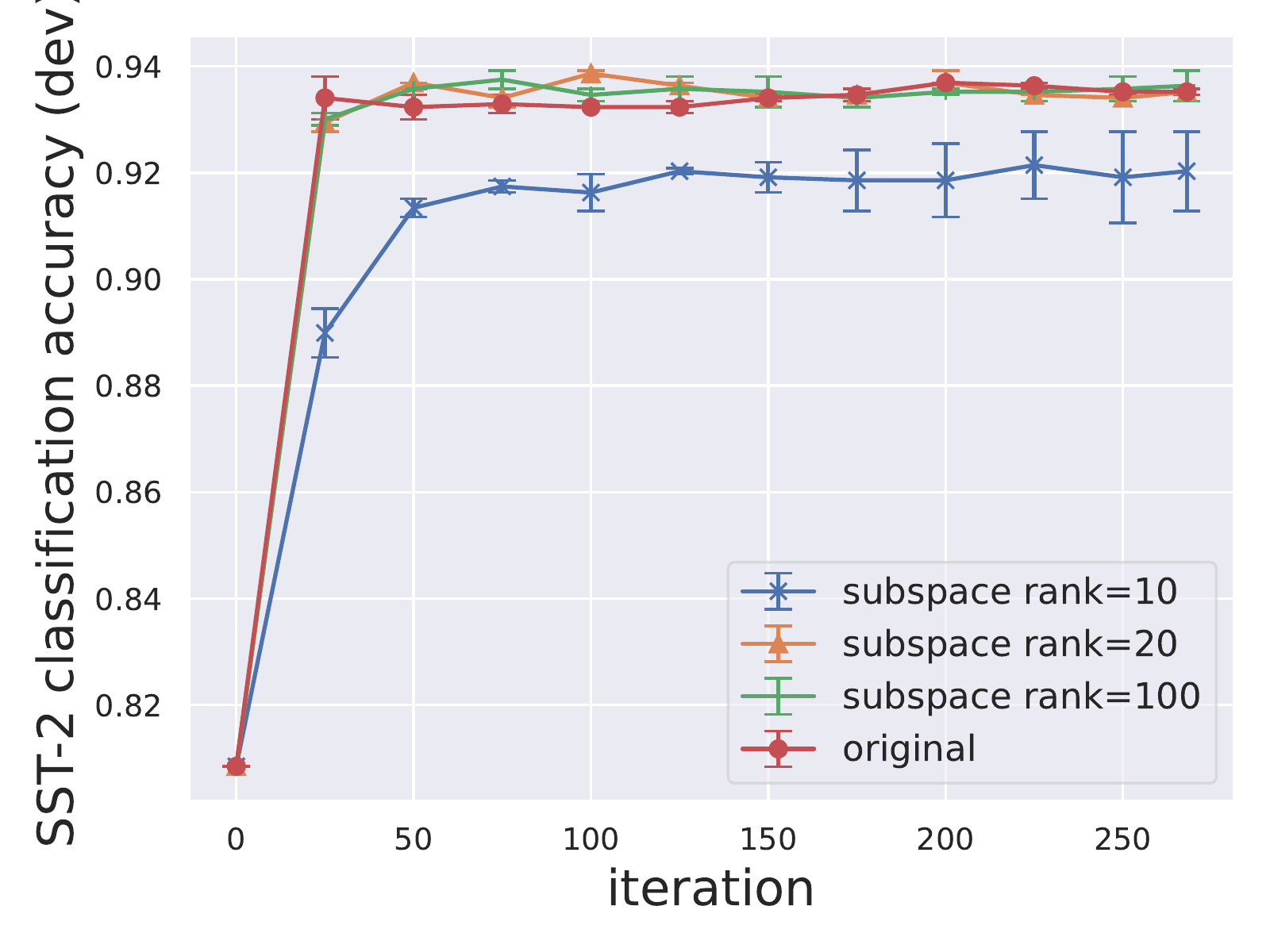}}
\footnotesize{(b) retrain in fixed subspace}
\end{minipage}
\end{center}
\caption{
Experiments for Roberta-large.
}
\label{fig:pca_roberta_large}
\end{figure}

\paragraph{Robustness to model size.}
In previous experiments, we empirically verified that gradients for fine-tuning DistilRoBERTa are near low rank.
Here, we show that similar observations also hold for Roberta-base and Roberta-large when fine-tuning only the attention layers.
The former setup has approximately $14$ million trainable parameters, while the latter has approximately $50$ million.
Figures~\ref{fig:pca_roberta_base} and~\ref{fig:pca_roberta_large} illustrate these results.

\paragraph{Non-robustness to fine-tuning strategy.}
Recall our fine-tuning experiments for classification was based on the template-completion formulation detailed in~\cite{li2021large}.
As opposed to framing the task as integer label prediction, this formulation requires the model to predict one of $K$ candidate tokens to fill in a templated prompt for a $K$-way classification problem.
While we have also performed the experiment where we re-train in the subspace of top principal components under the usual fine-tuning setup (stack a randomly initialized prediction head on top of the embedding of the \texttt{[CLS]} token), we found it difficult to recover the original fine-tuning performance when gradients are projected onto the top eigen-subspace with $d=100$ dimensions. 
Retraining performance exhibited high variance and the final dev accuracy was bi-modal over random seeds with near guess accuracy ($\approx50\%$) and original accuracy ($\approx90\%$) being the two modes.
We suspect this to be caused by the linear prediction head being randomly initialized.

\subsection{Additional Fine-Tuning Experiments with DistilGPT-2 for Generation Tasks}
\begin{figure}[thb]
\begin{center}
\begin{minipage}[t]{0.45\linewidth}
\centering
{\includegraphics[width=0.98\textwidth]{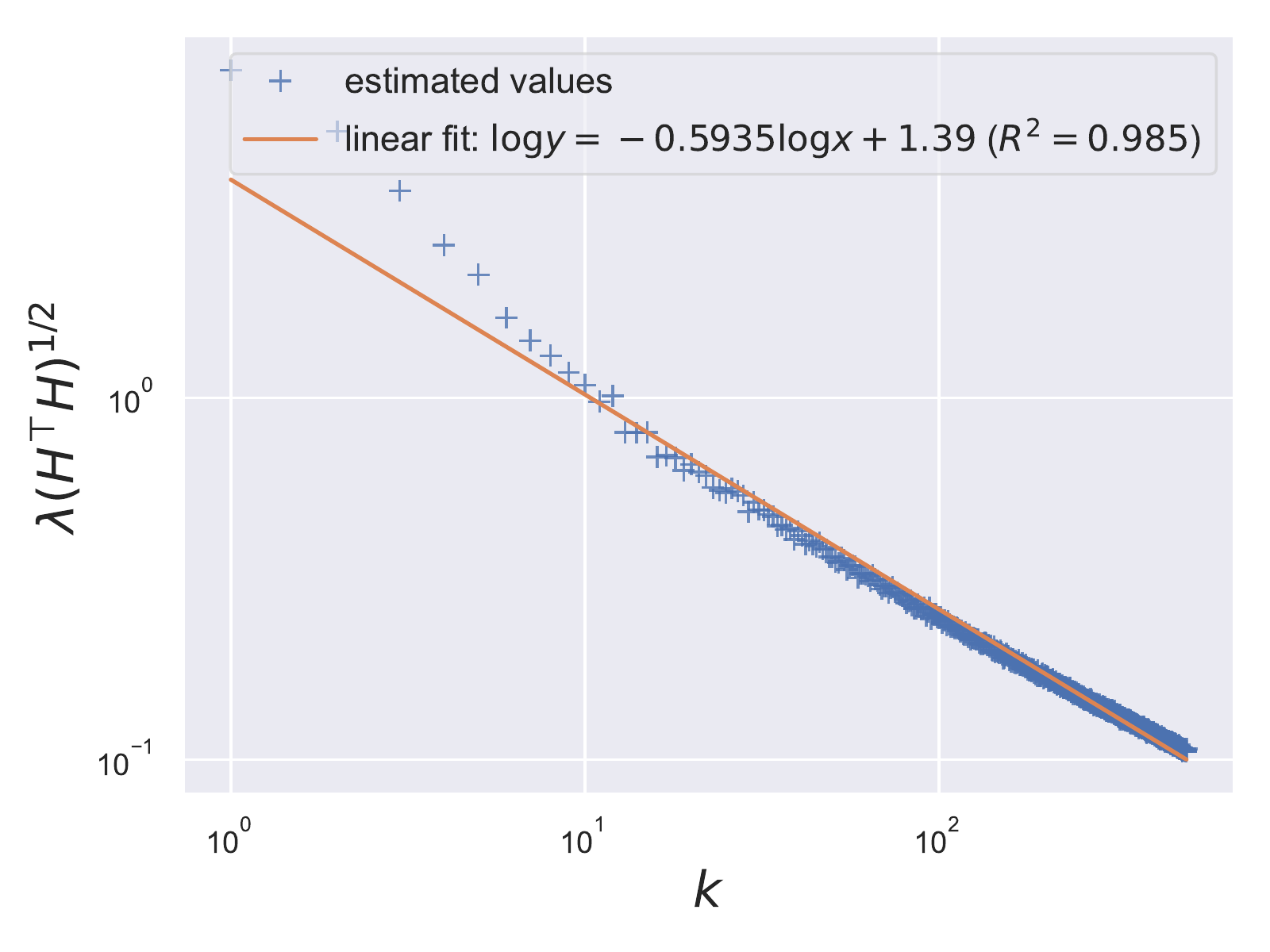}}
\end{minipage}
\end{center}
\caption{
The eigenspectrum of gradients from fine-tuning for text generation rapidly decays.
}
\label{fig:gpt2}
\end{figure}

Experiments in Section~\ref{sec:lm} demonstrated that gradients from fine-tuning for classification are mostly controlled by a few principal components.
In this section, we show that similar observations hold for fine-tuning on a text generation task. 
We follow the setup and hyperparameters in~\cite{li2021large} for privately fine-tuning DistilGPT-2 on the E2E dataset~\cite{novikova2017e2e} under $(8, \nicefrac{1}{n^{1.1}})$-DP. 
We fine-tune all weight matrices in attention layers that produce the queries, values, and keys. 
This amounts to fine-tuning approximately $10.6$ million parameters of a model with a total parameter count of more than $100$ million.
We again collected $r=4000$ gradients evaluated during private fine-tuning, performed PCA, and conducted the eigenspectrum analysis.
Figure~\ref{fig:gpt2} shows that the top eigenspectrum decays rapidly with a rate similar to what is observed in fine-tuning for the classification problem we studied.

\end{document}